# USING MACHINE LEARNING TO DEVELOP A NOVEL COVID-19 VULNERABILITY INDEX (C19VI)

**Anuj Tiwari [1], Arya V. Dadhania [2], Vijay Avin Balaji Ragunathrao [3], Edson R. A. Oliveira [4]**

[1] Department of Civil Engineering, Indian Institute of Technology, Roorkee, Uttarakhand, India; atiwari2@ce.iitr.ac.in
[2] College of Medicine, UIC, Chicago, Illinois, USA; adadha2@uic.edu
[3] Department of Pharmacology, UIC, Chicago, Illinois, USA; avin03@uic.edu
[4] Department of Microbiology and Immunology, UIC, Chicago, Illinois, USA; edsonrao@uic.edu

\* Correspondence: atiwari2@ce.iitr.ac.in; Tel.: +1 847 414 1752

## Summary

### Background

COVID-19 is now one of the most leading causes of death in the United States. Systemic health, social and economic disparities have put the minorities and economically poor communities at a higher risk than others. Every mitigation effort to contain the COVID-19 pandemic requires equitable resource allocation and region and situation-specific public health initiatives. Thus, there is an immediate requirement to develop a reliable measure of county-level vulnerabilities that can capture the heterogeneity of both vulnerable communities and the COVID-19 pandemic. This study reports a COVID-19 Vulnerability Index (C19VI) for identification and mapping of vulnerable counties in the United States.

### Methods

We proposed a Random Forest machine learning based COVID-19 vulnerability modeling technique using CDC's six themes (a) socioeconomic status, (b) household composition & disability, (c) minority status & language, (d) housing type & transportation, (e) epidemiological factors and (e) healthcare system factors. An innovative 'COVID-19 Impact Assessment' algorithm was also developed using homogeneity analysis (Pettitt's Test) and temporal trend assessment technique (Mann-Kendall and Sen's slope estimator) for evaluating severity of the pandemic in all counties and train the machine learning model. Developed C19VI was statistically validated and compared with the CDC COVID-19 Community Vulnerability Index (CCVI). Finally, using C19VI along with the census data, we explored racial inequalities and economic disparities in COVID-19 health outcomes amongst different regions in the United States.

### Findings

Our C19VI index indicates that 18.30% of the counties (575) falls into the 'very high' vulnerability class, 24.34% (765) in the 'high,' 23.32% (733) in the 'moderate', 22.34% (702) in the 'low,' and 11.68 % (367) in the 'very low.' The top ten states with the most percentage of 'very highly' and 'highly' vulnerable counties are Alabama (94%), Mississippi (90%), Louisiana (89%), Georgia (76%), South Carolina (76%), Arkansas (76%), Tennessee (76%), North Carolina (74%), Florida (70%) and New Mexico (70%). Furthermore, our index reveals that 75.57% of racial minorities and 82.84% of economically poor communities are very high or high COVID-19 vulnerable regions.

**Interpretation**

The proposed approach of vulnerability modeling is an innovative way to conduct COVID-19 vulnerability research, which takes advantage of both the well-established field of statistical analysis and the fast-evolving domain of machine learning. The C19VI provides an accurate and more reliable way to measure county level vulnerability in the United States. This vulnerability index aims at helping public health officials and disaster management agencies to develop more effective mitigation strategies especially for the disproportionately impacted communities in the United States.

## INTRODUCTION

The novel coronavirus disease (COVID-19) has been recognized as the newest and biggest global public health crisis[1]. In the first half of 2020, the COVID-19 has nearly killed half a million people worldwide of which more a quarter of them have died in the United States of America[2]. Within three months after the first reported case in the United States in January 2020, coronavirus cases had been confirmed in all fifty states, including the District of Columbia and other inhabited U.S. territories[2]. Shortly after, many states-imposed policies to curb the spread[3,4]. Despite these diffused efforts, a massive surge in incidence and mortality began to recur in June 2020 that placed the U.S. as the most affected country setting apart from other counties by a huge margin[2,5]. This has clearly indicated a lapse in effective COVID risk assessment and response at different levels. With the added concern of the second wave and disproportionate impact of the pandemic on minorities and economically poor, reliable country-wide assessment of COVID-19 vulnerability is a matter of necessity and urgency[6].

Identification of vulnerable areas is critical for public health departments to take the appropriate measures to increase preparedness against COVID-19. In order to identify these areas, the Centers for Disease Control and Prevention (CDC) initially used the social vulnerability index (SVI)[7], which is calculated based on census variables distributed in four distinct themes: i) socio-economic factors, ii) household composition, iii) minority status and language, and iv) access to housing and transportation. SVI failed to sufficiently determine vulnerability during this unprecedented situation, largely due to its primary objective aimed at addressing natural disaster crises during hurricanes, earthquakes, and forest fires[8,9]. Thus, the CDC and Surgo Foundation developed COVID-19 Community Vulnerability Index (CCVI)[10] by introducing two new variables, v) epidemiological risk factors and vi) public health system capacity, in the hope to rectify the shortcomings in the previous vulnerability assessment technique. Despite this optimization, CCVI and SVI are based on a statistical linear algorithm[7,10] that is unable to sufficiently account for the multiplicative, non-linear nature of vulnerability[11]. The lack of a comprehensive and accurate COVID-19 vulnerability index impairs the preparedness of critical areas against the pandemic as they are blurred to public health measures. This scenario highlights the urgent need for improvements of the nationwide approaches to identify vulnerable areas amid COVID-19 pandemic.

In the current study, we developed a more reliable assessment, the COVID-19 Vulnerability Index (C19VI) which quantifies the pandemic vulnerability of each United States county. This relative index processed the same six input variables as CCVI, however, instead of using a statistical linear algorithm, we utilized machine learning technique. We implemented Random Forest (RF) machine learning technique to calculate C19VI. An innovative 'COVID-19 Impact Assessment' algorithm was also developed using homogeneity analysis and temporal trend assessment techniques for training the RF model. Next, our C19VI index was compared with CDC's CCVI using advanced statistical measures and a machine learning model. We then tested the accuracy and checked the internal consistency of the C19VI.

Our vulnerability assessment methodology has allowed us to analyze the impact of COVID-19 that has been unequal yet widespread across the nation[12-15]. Besides, there are systemic socioeconomic inequalities that increase the susceptibility and exposure of the marginalized groups[13,15,16]. Thus, in addition to conducting a nationwide analysis of COVID-19 vulnerability,

C19VI has allowed us to explore the existing healthcare disparities in the realm of COVID-19 pandemic in great detail. This study may enhance the current techniques in vulnerability modeling, leveraging the preparedness of vulnerable counties to reduce the COVID-19 burden within the United States.

## MATERIAL AND METHODS

### INPUT DATASETS

We used publicly available datasets from Johns Hopkins University[2], Centers for Disease Control and Prevention (CDC)[10], United States Census Bureau[17], and United States Department of Homeland Security[18] for impact assessment, vulnerability modeling, population-specific vulnerability analysis, and data visualization and mapping, respectively. The data for COVID-19 confirmed cases and mortality in the United States from 22$^{nd}$ January 2020 – 31$^{st}$ July 2020 were obtained from Johns Hopkins University. Figure 1(A) and 1(B) presents the normalized (per 100,000) total confirmed cases and deaths dataset maps for all United States counties, respectively. The four socio-demographic SVI indicators and two CCVI thematic indicators, referred to as input themes were obtained from CDC. This results in six input themes that determine the vulnerability of a region to COVID-19 pandemic (Table 1). Figure 2 shows (A) socioeconomic status, (B) household composition & disability, (C) minority status & language, (D) housing type & transportation, (E) epidemiological factors and (F) healthcare system factors maps for the United States. County wise total population, racial population and poverty breakdown were obtained from the United States Census Bureau. Regional boundary data of the United States was collected from Homeland infrastructure foundation-level data[18] in Geographic Information System (GIS) ready file format (ESRI shapefile[19]). This study did not require a review by the Institutional Review Board since publicly available, de-identified data was used.

### COVID-19 IMPACT ASSESSMENT ALGORITHM

In order to understand the impact of COVID-19 pandemic in all 3142 counties in the United States, we have proposed a 'COVID-19 Impact Assessment' algorithm. This algorithm 'Scores' and 'Ranks' the impact of COVID-19 pandemic by evaluating the change with respect to time and severity in confirmed cases, deaths, and infection fatality rate (IFR)[20] using trend analysis (Mann Kendall[21,22] & Theil and Sen Slope[23,24]) and homogeneity assessment (Pettitt's test[25]). Trend analysis characterizes the overall pattern in daily time series dataset and homogeneity assessment identifies abrupt changes in temporal trends[21-25]. Thus, the algorithm classifies each county in one of the six impact groups, 'very high' (Rank = 1), 'high' (Rank = 2), 'moderate' (Rank = 3), 'low' (Rank = 4), 'very low' (Rank = 5) and 'non-significant' (Rank = -999). See supplementary material for the COVID-19 Impact Assessment Algorithm pseudocode.

**The algorithm functions in four steps:**

1. **Data import and pre-processing:** County-wise, daily time-series data of the confirmed cases and deaths were obtained from the John Hopkins University as mentioned above[2]. Then, daily time-series data for IFR is calculated using the imported datasets.
2. **Homogeneity analysis:** Pettitt's test[25] was applied county-wise to check for the homogeneity in the time-series dataset of all three epidemiological parameters obtained after step 1. If the data was found to be non-homogeneous, pre and post changepoint time series were computed and kept alongside the 'overall' dataset, which was the only populated data column in the cases of homogenous datasets. This expanded the time-series dataset into three aspects, i.e. pre changepoint, post changepoint, and overall, for each of the three epidemiological parameters, i.e. confirmed cases, deaths, and IFR for each county.
3. **Trend analysis:** We applied Mann Kendall's test[21,22] to assess the trend and its nature, i.e. increasing, decreasing, or no trend, in a given time-series. Next, the trend magnitude was quantified using the Theil and Sen slope estimator test[23,24]. Mann Kendall's, and Theil and Sen Slope estimator test was performed on all three time-series computed at the end of step 2 for all three epidemiological parameters in each county.
4. **COVID-19 Impact 'Score' and 'Rank' determination:** Impact Score was determined using the trend magnitude data obtained from the previous step. We used IFR as the most important parameter for assessing the impact of the COVID-19 pandemic in our algorithm[20,26]. In the instances where IFR did not show a significant trend in a given county, we first used the deaths[26]. If the deaths did not show a significant trend either, confirmed cases were used to evaluate the impact of the pandemic[26]. Thus, rank classification occurred in three stages, each further divided according to the homogeneity results:
    a. On the basis of the IFR:

i. In a homogeneous IFR time-series with an increasing 'overall' trend, the county was assigned Rank 1 and its impact Score was equal to the 'overall' trend magnitude.

ii. In a non-homogeneous IFR time-series with an increasing pre-changepoint trend, the scoring and ranking were specified based on the post-changepoint trend. Counties with increasing post-changepoint trends were classified as Rank 1, no post-changepoint trends as Rank 3, and decreasing post-changepoint trends as Rank 5. The Score of the counties with increasing (Rank 1) and no (Rank 3) post-change point trends were equal to the trend magnitude of 'post' and 'pre' time-series data, respectively. Finally, Score of the counties with decreasing post-changepoint data (Rank 5) was equal to the negative of the 'post' time-series trend magnitude.

b. On the basis of the fatalities (deaths):
   i. In a homogeneous death time-series with an increasing 'overall' trend, the county was assigned Rank 2 and its impact Score was equal to the 'overall' trend magnitude.

   ii. In a non-homogeneous death time-series with an increasing pre-changepoint trend, the scoring and ranking were specified based on the post-changepoint trend. Counties with increasing post-changepoint trends were classified as Rank 2, no post-changepoint trends as Rank 3, and decreasing post-changepoint trends as Rank 5. The Score of the counties with increasing (Rank 2) and no (Rank 3) post-changepoint trends were equal to the trend magnitude of 'post' and 'pre' time-series data, respectively. Finally, Score of the counties with decreasing post-changepoint data (Rank 5) was equal to the negative of the 'post' time-series trend magnitude.

c. On the basis of the confirmed cases:
   i. In a homogeneous time-series of confirmed cases with an increasing 'overall' trend, the county was assigned Rank 4 and its impact Score was equal to the 'overall' trend magnitude.

   ii. In a non-homogeneous time-series of confirmed cases with an increasing pre-changepoint trend, the scoring and ranking were specified based on the post-changepoint trend. Counties with increasing post-changepoint trends were classified as Rank 4, no post-changepoint trends as Rank 5, and decreasing post-changepoint trends as Rank 5. The Score of the counties with increasing (Rank 4) and no (Rank 5) post-change point trends were equal to the trend magnitude of 'post' and 'pre' time-series data, respectively. Finally, Score of the counties with decreasing post-changepoint data (Rank 5) was equal to the negative of the post time-series trend magnitude.

Every other county was classified as Rank -999 and Score -999. Finally, out of the three ranks, assigned to each county, based on the three epidemiological variables, the highest impact group (lowest rank) and its corresponding trend magnitude were decided as the final COVID-19 Impact Score and Rank for a given county.

**GENERATION OF COVID-19 VULNERABILITY INDEX (C19VI)**

Our study methodology was built and tested in six steps (Figure 3). First, the training-testing data was prepared using the "most affected" and the "non-significantly" affected counties using the proposed COVID-19 impact assessment algorithm. Second, COVID-19 vulnerability map was generated using the RF machine learning technique[27,28]. Third, vulnerability modeling was validated using Receiver Operating Characteristic (ROC)-Area Under the ROC Curve (AUC) technique[29-31] and Cronbach's $\alpha$[32]. Fourth, our C19VI modeling was comparatively assessed against the CDC's CCVI using Friedman[33] and two-tailed Wilcoxon signed rank[34] test and later, the input themes contribution to the respective vulnerability index, the output, were ranked using, and Boruta technique[35]. Fifth, C19VI was analyzed with racial minority population and poverty dataset to determine the disproportionate impact of COVID-19 pandemic on county level. Lastly, an interactive version of the C19VI map with other results was released to the public using the ESRI Web GIS customization toolkit[36]. Each step is further detailed below:

**Step 1: Preparation of the training-testing dataset:** Proposed COVID-19 Impact assessment algorithm was used to map the impact of COVID-19 pandemic on all 3142 counties in the US using confirmed cases and deaths. Out of total 3142 counties, 200 very highly affected and 200 non-significantly affected counties were selected to prepare the COVID-19 vulnerability modeling training and testing dataset. 70% of the total counties (280) were randomly selected and implemented as a training dataset while rest 30% (120) were used for testing.

**Step 2: COVID-19 vulnerability modeling:** COVID-19 vulnerability modeling was implemented using the RF machine learning technique[27,28]. This model predicts vulnerability of a given county on a continuous scale of 0 (least vulnerable) to 1 (most vulnerable). The map was graded according to the COVID-19 vulnerability index into five vulnerability classes[10]: very high (>80%), high (80%-60%), moderate (60%-40%), low (40%-20%) and very low (<20%)[10].

**Step 3: Validation of vulnerability modeling:** The effectiveness of RF machine learning technique was specified by evaluating uncertainties in the resulting vulnerability map. The ROC - AUC[29-31] was implemented to analyze the conformity between the validation fold of the training-testing dataset and the products of the applied technique. We computed Cronbach's $\alpha$[32] for the developed C19VI index to measure reliability by assessing the C19VI values, the output, with CDC's six theme variables, the input.

**Step 4: Comparison of the CCVI and C19VI:** As, both the CCVI and the C19VI models are developed using the same six thematic indicators, Friedman[33] and two-tailed Wilcoxon signed rank[34] statistical tests were implemented to comparatively assess model vulnerability prediction ability. Next, Boruta feature importance assessment technique[35] was used to evaluate the relative importance of input indicators in CCVI and C19VI.

**Step 5: Community specific vulnerability analysis:** Long-standing systemic, social and economic inequities across the counties have put many people from racial minority groups and living below the poverty line at increased risk of getting sick and dying from COVID-19[15,16,37]. By overlaying the C19VI map on racial minority population percentage data, COVID-19 vulnerability specific to racial minority groups were identified. As recommended by CDC, a 13% of the racial minority threshold was used for computing the COVID-19 vulnerability for racial minority groups[38]. Similarly, by overlaying the C19VI map on poverty percentage data, COVID-19 vulnerability specific to economically poor communities were identified. As defined by the Economic Research Service (ERS), USDA a 20% of the poverty threshold was used to estimate the vulnerability for economically poor communities[39,40]. ESRI ArcGIS overlay analysis tool[41] was used to conduct the community-specific vulnerability analysis.

**Step 6: Customization of C19VI Web Map Viewer:** ESRI Web App Builder[36] was used to develop an interactive 'C19VI Web Map' portal. This portal features three layers: 'C19VI' layer, 'COVID-19 Vulnerability (Racial Minority) - C19VI' layer, and 'COVID-19 Vulnerability (Poverty) - C19VI' layer. Every layer features its own attributes. The 'C19VI' layer displays the C19VI values, COVID-19 Impact Rank, total number of confirmed cases and deaths as of July 31st 2020 for each United States county. The 'COVID-19 Vulnerability (Racial Minority) - C19VI' layer displays the C19VI and minority population percentage for each United States county. The 'COVID-19 Vulnerability (Poverty) - C19VI' layer displays the C19VI and poverty percentage for each United States county.

## RESULTS

### COVID-19 IMPACT ASSESSMENT

Our COVID-19 Impact Assessment algorithm performed a county-wise assessment of the pandemic using the confirmed cases, deaths and IFRs data from 22nd January 2020 to 31st July 2020. We generated a map of our assessment that groups the impact of the pandemic on all United States counties in one of the six categories (Figure 4(A)). We found 88 counties with 'very high', 30 with 'high', 73 with 'moderate', 344 with 'low' and 214 with 'very low,' and 2393 with 'non-significant' impact due to the COVID-19 pandemic (Figure 4(B)). Top 200 counties with the most impact and the bottom 200 with non-significant impact were used as training and testing datasets for our COVID-19 vulnerability model.

### COVID-19 VULNERABILITY MODELING

Using the impact assessment data of the selected United States counties, input themes and the RF technique, we developed COVID-19 Vulnerability Index (C19VI). Figure 5(A) shows the C19VI map at the scale of 0 to 1. As presented in Figure 5(B), we computed C19VI for all United States counties and classified them in one of the five vulnerability Classes, 'very high', 'high', 'moderate', 'low', and 'very low'. We found that 11.68% of the counties (367) fall into the 'very low' category, 22.34% (702) in the 'low,' 23.32% (733) in the 'moderate,' 24.34% (765) in the 'high,' and 18.30% (575) in the 'very high' category (Figure 5(C)). Based on C19VI values, 20 most and least vulnerable counties and their corresponding input theme contribution to the vulnerability are displayed alongside CDC's CCVI in Figure 6(A) and Figure 6(B), respectively.

## MODEL VALIDATION AND RELIABILITY

We used the AUC-ROC technique to validate the prediction accuracy of our C19VI model. As shown in Figure 7(A) and 7(B), we found 90% accuracy (AUC = 0.90) during the training phase and 84% accuracy (AUC = 0.84) during the testing phase, respectively. High internal consistency (Cronbach's α = 0.709) of C19VI model was revealed using Cronbach's α test[42,43]. Overall, validation and reliability results indicate that the random forest machine learning modeling provides a high quality COVID-19 vulnerability map (C19VI) for the United States[31,42,43].

## COMPARATIVE EVALUATION – CCVI & C19VI

Since we introduced a new COVID-19 vulnerability assessment index, we quantitatively evaluated its performance against the existing vulnerability model, CDC's CCVI, to assess C19VI's predictive power and applicability. We used Friedman test[33], two-tailed Wilcoxon signed rank test[34], and Boruta parameter importance assessment technique[35] to comparatively evaluate C19VI and CCVI.

- **Friedman and Wilcoxon tests:** The Friedman and the two-tailed Wilcoxon signed rank tests detected significant differences (p<0.0001) between the indices given by the two models, C19VI and CCVI. The mean rank of C19VI is 1.614 while mean rank of CCVI is 1.386. The full results of the Friedman and Wilcoxon tests can be found in Table 2 and Table 3, respectively.

- **Boruta Test:** Individual importance of each CDC input themes in determining C19VI and CCVI were quantified using Boruta, a wrapper algorithm. The most important parameter in C19VI was theme 5 (135.17), while in CCVI, it was theme 6 (131.36) (Table 4). Moreover, theme 3 (Minority Status & Language) ranked second in parameter importance in the C19VI model as compared to CCVI, where it is ranked fourth (Table 4). The full results of the Boruta test on CCVI and C19VI are presented in Figure 8(A) and 8(B), respectively.

## COMMUNITY SPECIFIC VULNERABILITY ANALYSIS

The racial minority populations of the United States reside more densely in the southern states and in urban areas[17,44,45]. Our community-specific analysis reveals that the racial minorities disproportionately live in counties that are more vulnerable to COVID-19 (Figure 9(A)). We found that 75.57% counties with racial minority populations > 13%, have very high or high (CCVI > 0.60) COVID19 vulnerability. Similar to racial minorities, economically poor communities are more likely to be affected by the virus and have higher mortality rates[45]. The C19VI derived COVID-19 vulnerability with reference to poverty is presented in Figure 9(B). We find that 82.84% of economically poor counties, where poverty > 20%, have very high or high (CCVI > 0.60) COVID-19 vulnerability.

**C19VI WEB MAP VIEWER**

The Urban Data Visualization Lab (UDVL) at the University of Illinois at Chicago (UIC) have featured an interactive version of the C19VI map which can be accessed at the following URL https://udv.lab.uic.edu/national-covid-19-vulnerability-index-c19vi. This map portal is easily accessible on personal computers and mobile devices. Figure 10(A) is a snapshot from the web portal illustrating the variation in C19VI across the nation and the pop-up widget displaying a tabular view of C19VI attributes. Similarly, Figure 10(B) and Figure 10(C) depicts the 'COVID-19 Vulnerability (Racial Minority) - C19VI' and 'COVID-19 Vulnerability (Poverty) - C19VI' layers with their corresponding attributes, respectively.

**DISCUSSION**

Ever since the United States declared a national emergency due to the COVID-9 pandemic in March 2020, the country is grappling against a huge continuous surge in the incidence and mortality rates[2,46]. In the last month alone, June 30th, 2020 to July 31st, 2020, the total number of confirmed cases has risen from 27,29,764 to 47,13,014 while the total death count has increased from 1,30,313 to 1,56,826[2]. It is expected that the United States' COVID-19 death toll will double, potentially reaching more than 0.4 million by the beginning of next year[6]. Thus, keeping in mind the uncontrollable spread, ineffective strategies to check the transmission, disproportionate impact based on systemic inequalities, and heterogeneous impact on different regions, we have developed a county-level COVID-19 vulnerability index (C19VI) that assess the vulnerability of a region using an innovative methodology. This methodology takes into account the limitations of the existing vulnerability modeling techniques to assess COVID-19 vulnerability nationwide and performs a disproportionate analysis that points out the existing health disparities in the country.

In the following sections, we discuss the unique characteristics of our C19VI model and the utility of C19VI in the nationwide and community-specific vulnerability assessment.

- **NOVEL APPROACH OF VULNERABILITY MODELING**

    Recently, many researchers[47-54] have internationally conducted COVID-19 risk and vulnerability assessment using either mathematical modeling[48] and linear statistical techniques with sociodemographic, economic, and health indicators, for example, equal weight assignment[10,49], principal component analysis (PCA)[50,51], or heuristic modeling[52], or have performed numerical simulations of the total confirmed cases, deaths, and IFRs using statistical[47,53] and machine learning[54] techniques to compute COVID-19 specific vulnerability. While these approaches have enhanced the domain of pandemic vulnerability modeling, there is at least one of the three underlying limitations that prevent optimum modeling. Either they implement an equal weight assignment approach in vulnerability assessment, assume steady transmission rates in mathematical modeling, or treat confirmed cases, deaths, and IFR as constants for vulnerability assessment. However, it is known that 1) not all input themes variables are equally important in determining vulnerability[49], 2) confirmed cases, deaths, and IFR are not biological constants in a pandemic and thus, they do reflect the severity of the pandemic in a particular context, at a particular time[55], 3) selection of constant pandemic transmission rates for both mathematical analysis and data fitting is unrealistic in nature and does not encounter the implications of government implemented disease control actions and individuals' voluntary responses against COVID-19[55,56].

    Thus, we optimized these limiting factors by introducing RF machine learning, a non-linear, non-parametric predictive modeling technique which can efficiently compute large datasets and account for complex interactions between variables[27,28]. Additionally, in comparison to other machine learning techniques, RF is easier to tune and does not overfit the data making it ideal for pandemic vulnerability modeling[28].

    Furthermore, we optimized dynamic characteristics of a pandemic by developing novel 'COVID-19 Impact Assessment' algorithm which assesses the regional pandemic impact by performing trend and homogeneity analysis on daily datasets rather than static values for a defined period. Trend and homogeneity assessments help characterize the

course of the pandemic and point out the COVID-19 response through changes in healthcare infrastructure or policies in a given region by identifying subtle changes in daily datasets[21-25]. Moreover, besides optimization, our impact assessment algorithm also serves to enhance vulnerability modeling to be driven by the chronic disease burden, healthcare infrastructure, and policy impact such as lockdown phases.

In conjunction with the optimized impact assessment algorithm, high training (90%) and testing (84%) accuracy with favorable internal reliability score (Cronbach's α = 0.709) of the RF machine learning-derived predictive modeling technique makes C19VI an accurate and reliable index. Besides, despite using the same input, the machine learning-derived C19VI produced significantly different and consistent results than the CDC's CCVI as elucidated through the Friedman and Wilcoxon signed-rank tests. Moreover, Boruta algorithm-based importance assessment of the variables for both methods shows that both methods handled the variables with major notable differences. The divergence between the two methods indicates that the C19VI was able to capture non-linear relationships in the variables which were not captured with the linear 'equal weight assignment approach' used in the CDC's CCVI model.

Therefore, capturing non-linearity in the input variables, promising validation results, and unique characteristics of the C19VI methodology makes it an optimal candidate to be incorporated into more comprehensive, quantitative vulnerability assessment methodologies.

- **NATIONWIDE VULNERABILITY ANALYSIS**

Our nationwide vulnerability analysis reveals interesting patterns of vulnerability distributions around the country. We found that the majority of most vulnerable counties are concentrated in the southern states. As shown in the Figure 5(B), nine of the top ten states with the most percentage of 'very highly' and 'highly' vulnerable counties—Alabama (94%), Mississippi (90%), Louisiana (89%), Georgia (76%), South Carolina (76%), Arkansas (76%), Tennessee (76%), North Carolina (74%), Florida (70%) and New Mexico (70%)—are southern states. Secondly, although the counties in the northeastern states had significant confirmed cases, many states in this region have low vulnerability. Four of the top ten states with the highest percentage of 'very low' and 'low' vulnerable counties—Connecticut (100%), New Hampshire (100%), Wyoming (96%), North Dakota

(91%), Wisconsin (83%), Nebraska (83%), Montana (82%), Maine (81%), Vermont (79%), Minnesota (78%)—are northeastern states. Thus, we see that the counties with high and low vulnerability are clustered together, respectively, and the similar vulnerability classes are distributed discretely in the different geographical regions of the United States. This non-uniform, region-dependent distribution of COVID-19 vulnerability in the United States can be associated with adoption of different public health strategies on a state-level and with the regional sociodemographic distribution in the United States.

This index can also be used alongside other epidemiological data, such as disease transmission, infection fatality rate, the proportion of cases needing hospitalization, intensive care unit admissions, or ventilator support to heighten the preparedness of a district or state, as well as planning and executing the response. We also recommend the use of our C19VI index alongside the CDC's Social Vulnerability Index (SVI) for developing disaster risk assessment and preparedness plans in COVID-19 affected regions. For example, in the times of COVID-19 pandemic, the C19VI should be used alongside the SVI for the disaster management in counties with frequent forest fires, tornadoes or hurricanes.

- **RACIAL/ECONOMIC DISPROPORTIONALITY USING VULNERABILITY**

    COVID-19 has brought previously unaddressed health disparities of racially marginalized and economically poor communities to the forefront of both disaster management officials and government concern. By overlaying the C19VI with the race and poverty data, we find that racial minorities and economically poor Americans disproportionately live in communities that are more vulnerable to COVID-19. Our finding is consistent with other evidence highlighting the disproportionate incidence of COVID-19 among minority groups and poor communities[13,15,37,57-59]. In an ongoing pandemic, when the available county-level cases and deaths dataset segregated by minority population and economic status are not sufficient to generate reliable COVID-19 risk estimates, our analysis provides an excellent way to identify precise solutions, from neighborhood-level investments in cities to targeted testing strategies in rural areas, to help the communities that disproportionately bear the burden of this crisis.

Thus, the C19VI is intended to help policymakers at all levels, non-profit entities, private companies, local organizations, and the general public to help in effective COVID-19 contingency planning, drive the equitable distribution of resources, address pandemic-associated healthcare disparities, provide businesses with opportunities to grow where support is needed the most, raise public awareness on COVID-19 pandemic for informed decision making. Besides, we hope that this methodology will also prove to be useful in driving more advanced predictive modeling techniques by professionals in academia and innovation. Lastly, we believe that C19VI will be a tool for public health officials to help formulate effective and inclusive COVID response plans that will alter the course of this pandemic.

**LIMITATIONS**

Ideally, it would be possible to calculate the index at a census tract level. However, several important variables used to define vulnerability were not available at a census tract level. Hence, this analysis is restricted to the county-level. Secondly, being based on the ranking of counties for CDC six themes, our C19VI is a relative index of each county rather than being an absolute score. Thirdly, we were unable to test the external validity of C19VI since no accurate and stable measure of vulnerability was available. Finally, confirmed cases and deaths used in this study are up to 31st July 2020 and might not have captured the impact of the pandemic as well as the strength of the counties in which rapid changes have occurred up to the present day.

**CONCLUSION**

In this work, we proposed an innovative approach to conduct the vulnerability assessment of COVID-19 within the United States at the county level. This approach integrates the reliable and high-functioning domains of machine learning and predictive modeling. RF machine learning technique in conjunction with novel 'COVID-19 Impact Assessment' algorithm successfully yields a vulnerability model, which non-linearly processes the input variables with high training-testing accuracy and favorable internal validity, that generates a COVID-19 vulnerability index of the United States. Besides promising validation results, comparative assessment of our technique confirms that the C19VI predictive model is a reliable and pragmatic alternative to the CDC's CCVI. More importantly, our innovative approach to develop a vulnerability index has enhanced the nationwide capacity to predict the potential harms accurately which may help curtail the distressing course and consequences of the pandemic. Additionally, when combined with the racial minority and poverty dataset, C19VI demonstrated its efficiency in disproportionate vulnerability analysis and helped validate the existing disparities. Thus, as a COVID-19 risk assessment tool, this index will help the public health officials of the federal, state, and local agencies and other academic and autonomous institutions in formulating and implementing policies, plans, and recommendations. Lastly, further exploration of our methodology, especially in vulnerability modeling, has the potential to progress the digital public health technology for pandemic management in general.

**CONTRIBUTORS**
AT conceived the study, designed methodology and programmed the model. AT and VA-BR performed statistical analysis and prepared the figures. AVD worked on the data acquisition and data interpretation. AT, AVD and ERAO performed data analysis. AT and AVD wrote the manuscript. AVD and ERAO revised the manuscript critically for important intellectual content. All authors interpreted the results, and approved the final version for submission.

**DECLARATION OF INTERESTS**
We declare no competing interests.

**ACKNOWLEDGEMENTS**
We would like to thank Johns Hopkins University, Centers for Disease Control and Prevention (CDC), United States Census Bureau and Department of Homeland Security for sharing information necessary for the outbreak investigation, vulnerability modeling, analysis and mapping. We sincerely acknowledge the support and collaboration of Dr. Moira Zellner, Director, Urban Data Visualization Lab (UDVL), CUPPA, UIC, Chicago, USA. We thank Anton Rozhkov, Ph.D. Candidate, Urban Data Visualization Lab (UDVL), CUPPA, UIC, Chicago, USA for customizing the web map portal and Abhilasha Dixit, Research Scholar, Centre of Excellence in Disaster Mitigation and Management (CoEDMM), Indian Institute of Technology-Roorkee, India for her expert technical assistance.

**DATA SHARING**
We used publicly available data and have referenced the sources in the paper.

# TABLE

**Table 1. The CDC's CCVI theme indicators and corresponding variables.**

| Theme | Indicator | Type | Variable | Resolution |
|---|---|---|---|---|
| 1 | Socioeconomic Status | Social | Below Poverty | Census Tract |
| | | | Unemployed | Census Tract |
| | | | Income | Census Tract |
| | | | No High School Diploma | Census Tract |
| 2 | Household Composition & Disability | Social | Aged 65 or Older | Census Tract |
| | | | Aged 17 or Younger | Census Tract |
| | | | Older than Age 5 with a Disability | Census Tract |
| 3 | Minority Status & Language | Social | Minority | Census Tract |
| | | | Speaks English "Less than Well" | Census Tract |
| 4 | Housing Type & Transportation | Social | Multi-Unit Structures | Census Tract |
| | | | Mobile Homes | Census Tract |
| | | | Crowding | Census Tract |
| | | | No Vehicle | Census Tract |
| 5 | Epidemiological Factors | COVID | Cardiovascular Conditions | County |
| | | | Respiratory Conditions | County |
| | | | Immuno- compromised | County |
| | | | Obesity | County |
| | | | Diabetes | County |
| | | | Population Density | Census Tract |

|   |   |   | Influenza and Pneumonia Death Rates | County |
|---|---|---|---|---|
| 6 | Healthcare System Factors | COVID | Health System Capacity | State/ Hospital Region |
|   |   |   | Health System Strength | State/ County |
|   |   |   | Health System Preparedness | State/ County |

**Table 2. Results of Friedman test for CCVI and C19VI.**

| Index | Degrees of Freedom | Chi-Squared value | p-value | Mean Rank |
|---|---|---|---|---|
| C19VI | 1 | 3.841 | < 0.0001 | 1.614 |
| CCVI |   |   |   | 1.386 |

**Table 3. Comparison of CCVI and C19VI using two-tailed Wilcoxon signed-rank test.**

| Pairwise Comparison | z-Statistic | p-value |
|---|---|---|
| CCVI – C19VI | -12.461 | < 0.0001 |

**Table 4. Importance assessment of the input themes in CCVI and C19VI using Boruta algorithm.**

| Theme | Indicator | CCVI - Mean importance | C19VI - Mean importance |
|---|---|---|---|
| 1 | Socioeconomic Status | 74.56 | 105.60 |
| 2 | Household Composition & Disability | 43.20 | 34.45 |
| 3 | Minority Status & Language | 68.69 | 112.98 |
| 4 | Housing Type & Transportation | 47.81 | 67.86 |
| 5 | Epidemiological Factors | 102.84 | 135.17 |
| 6 | Healthcare System Factors | 131.36 | 72.25 |

**FIGURES**

**Figure 1 Maps of the U.S. counties representing confirmed COVID-19 cases and deaths normalized (per 100,000 people) upto 31st July 2020.** (A) County-wise confirmed cases in the U.S. (B) County-wise deaths in the United States.

**Figure 2 Maps of the U.S. counties representing the CDC's six COVID-19-specific input themes.** Panels A-F shows the CDC's scoring of the respective sociodemographic variables in the U.S. (A) Socioeconomic Status, (B) Household Composition & Disability, (C) Minority Status & Language, (D) Housing Type & Transportation, (E) Epidemiological Factors and (F) Healthcare System Factors maps for the United States.

**Figure 3 Flow diagram of the C19VI methodology.** The figure illustrates the C19VI methodology, from input datasets to customization of the C19VI Web Map Viewer, in step-wise details.

**Figure 4 COVID-19 Impact Assessment algorithm output.** (A) Map of COVID-19 impact showing classification of each U.S. county in one of the six impact ranks generated using COVID-19 Impact Assessment Algorithm. The map shows counties with very high impact (Rank 1) in yellow, high (Rank 2) in orange, moderate (Rank 3) in red, low (Rank 4) in rose, very low (Rank 5) in purple, and insignificant (Rank 999) in tropical blue. (B) Bar graph depicting total number of counties in each COVID-19 Impact Rank.

**Figure 5 COVID-19 vulnerability modeling output.** (A) COVID-19 Vulnerability Index (C19VI) map on a continuous scale of 0 (least vulnerable) to 1 (most vulnerable). (B) Map of C19VI showing classification of each U.S. county in of the six vulnerability classes generated using the RF machine learning-derived C19VI model. The map shows counties with very high vulnerability (C19VI 0.8-1.0) in red, high (0.6-0.8) in vermillion, moderate (0.4-0.6) in orange, low (0.2-0.4) in amber, and ver y low (0.2-0.0) in yellow. (C) Bar graph depicting the total number of counties in each C19VI class.

**Figure 6 Heat map of CCVI and C19VI comparative assessment.** (A) Heat map of CCVI and C19VI alongside the six input theme indicators of the twenty most vulnerable counties shortlisted based on the C19VI. (B) Heat map of CCVI and C19VI alongside the six input theme indicators of the twenty least vulnerable counties shortlisted based on the C19VI.

**Figure 7 C19VI model validation using ROC-AUC technique.** (A) ROC-AUC curve with 90% accuracy during the training phase. (B) ROC-AUC curve with 84% accuracy during the testing phase.

**Figure 8 Importance assessment of the six input theme indicators in CCVI and C19VI using Boruta algorithm.** Panel A and B shows the box plot summary of the Boruta input parameter importance assessment of CCVI and C19VI, respectively. (A) Box plot (mean, median, minimum, and maximum Z) for each input theme in the increasing order of importance on the X-axis for the CCVI model. (B) Box plot (mean, median, minimum, and maximum Z) for each input theme in the increasing order of importance on the X-axis for the C19VI model.

**Figure 9 C19VI facilitated community-specific vulnerability assessment.** C19VI of all U.S. counties was overlaid with racial minority population percentage data and poverty percentage data to generate panel A and B, respectively. (A) Map of the U.S. counties showing COVID-19 vulnerability of the racial minorities. The map shows counties with high vulnerability (C19VI > 0.6) and higher than 13% racial minorities in cobalt, low vulnerability (C19VI < 0.6) and higher than 13% racial minorities in tropical blue, high vulnerability (C19VI > 0.6) and lower than 13% racial minorities in red, and low vulnerability (C19VI < 0.6) and lower than 13% racial minorities in chardonnay. (B) Map of the U.S. counties showing COVID-19 vulnerability of the economically poor communities. The map shows counties with high vulnerability (C19VI > 0.6) and higher than 20% poverty in red, low vulnerability (C19VI < 0.6) and higher than 20% poverty in pink, high vulnerability (C19VI > 0.6) and lower than 20% racial minorities in orange, and low vulnerability (C19VI < 0.6) and lower than 20% poverty in chardonnay.

**Figure 10 COVID-19 Web Map Viewer.** This figure illustrates snapshots of the COVID-19 Web Map Viewer, which can be accessed at https://udv.lab.uic.edu/national-covid-19-vulnerability-

index-c19vi. (A) Distribution map of 'C19VI' layer across the nation and the pop-up widget displaying a tabular view of C19VI value, COVID-19 Impact class, confirmed cases, and deaths for the selected county. (B) Distribution map of 'COVID-19 Vulnerability (Racial Minority) - C19VI' layer and the pop-up widget displaying a tabular view of C19VI value and racial minority percentage for the selected county. (C) Distribution map of 'COVID-19 Vulnerability (Poverty) - C19VI' layer and the pop-up widget displaying a tabular view of C19VI value and poverty percentage for the selected county.

**Figure 1**

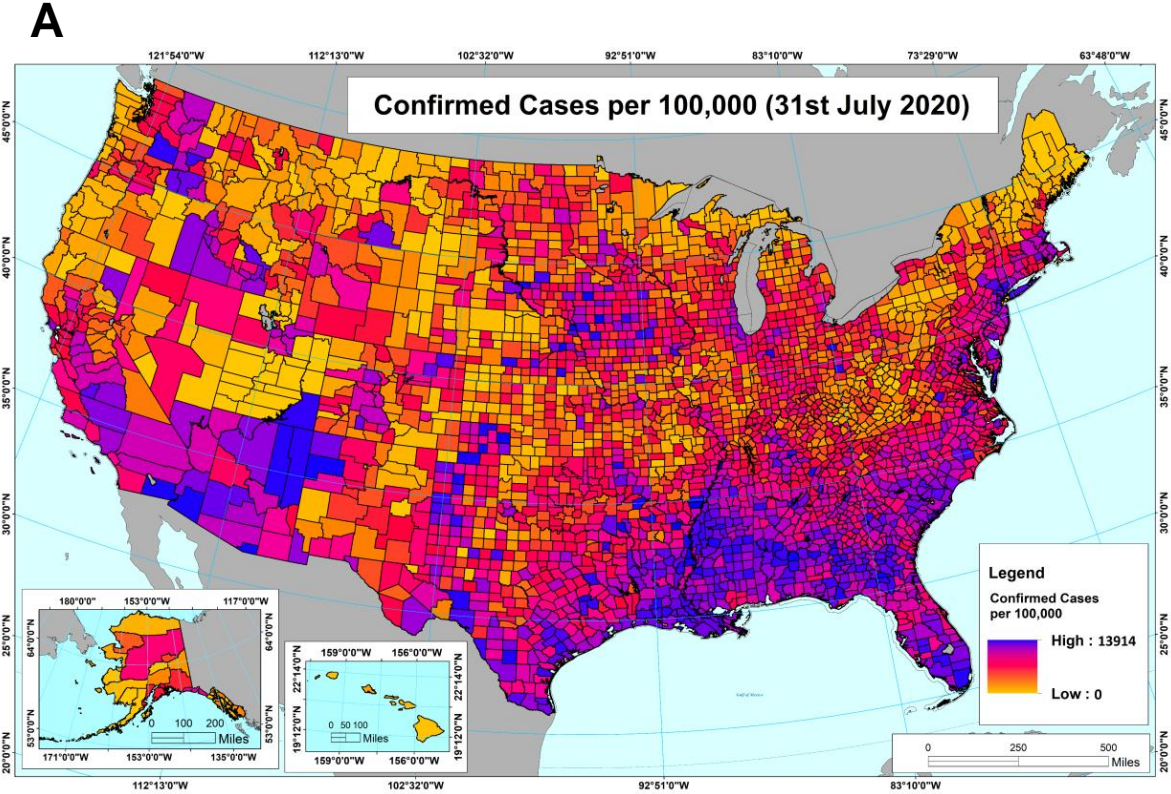 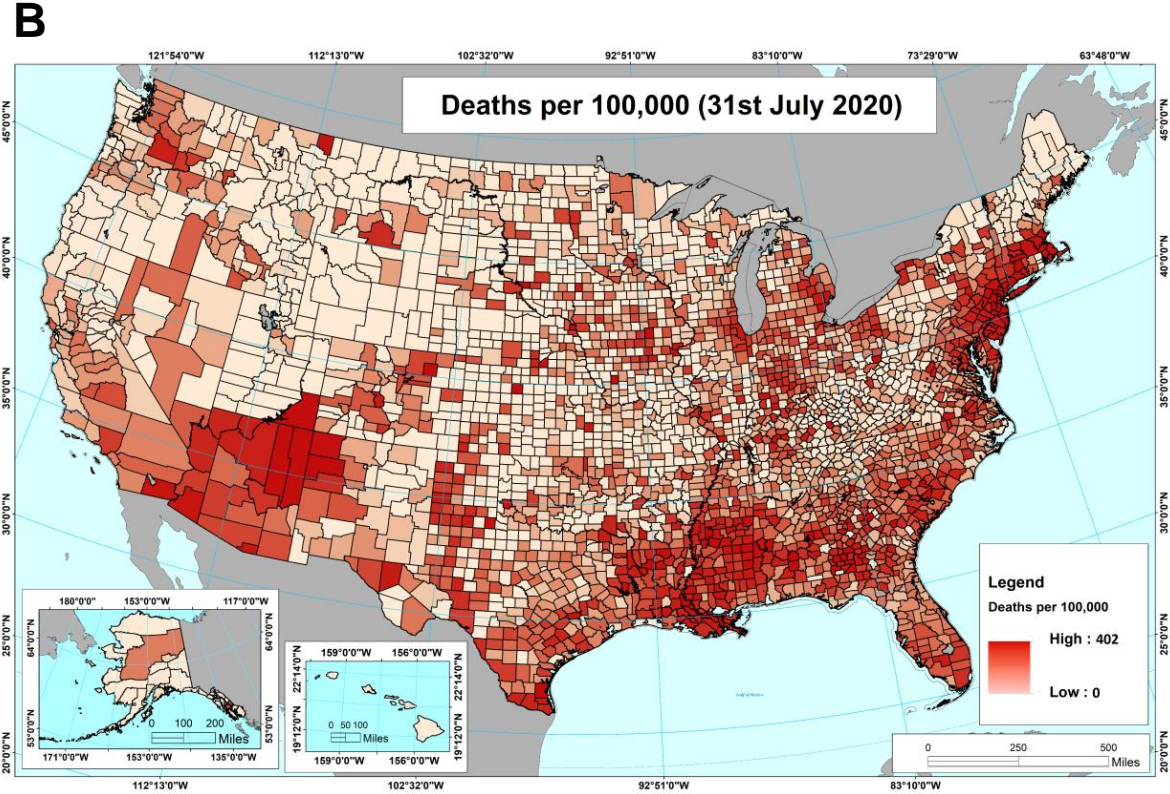

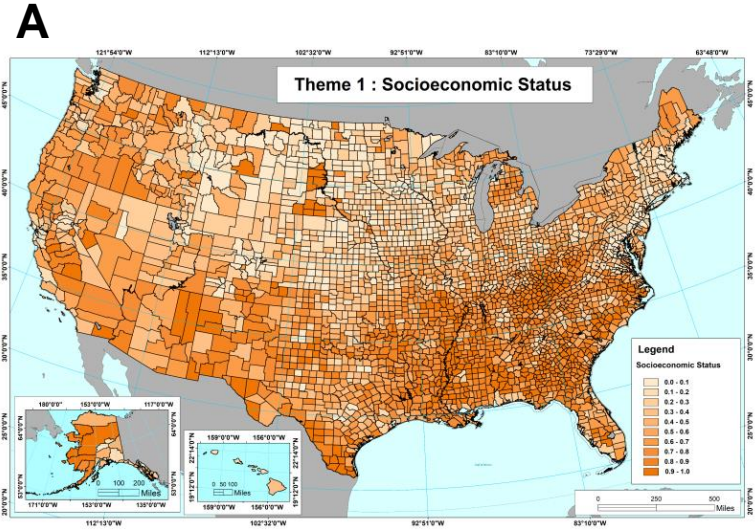
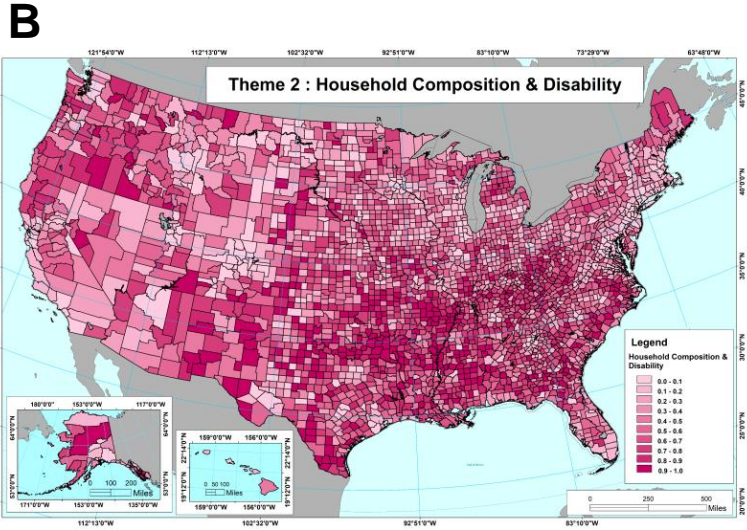
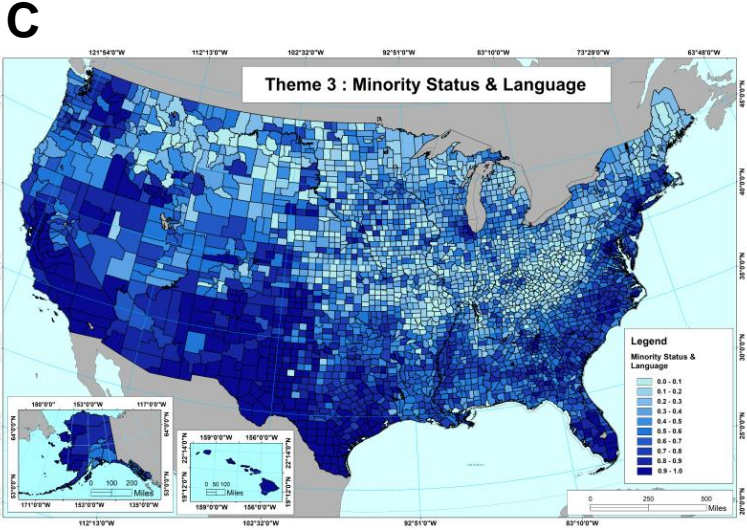
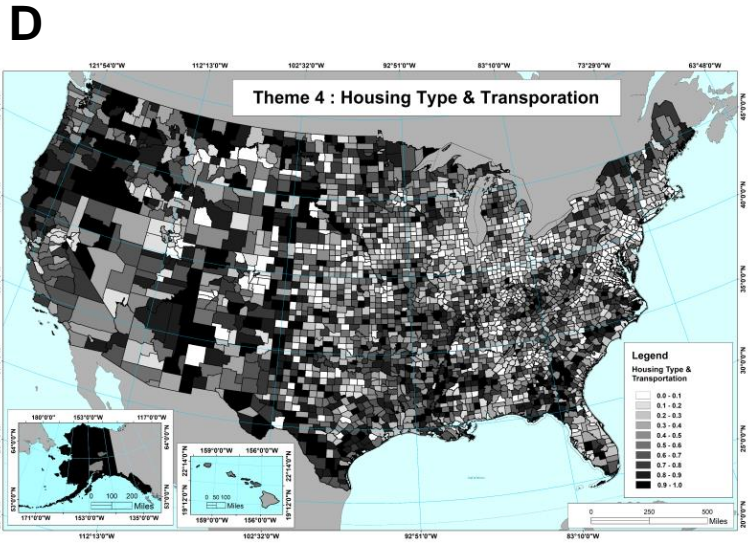
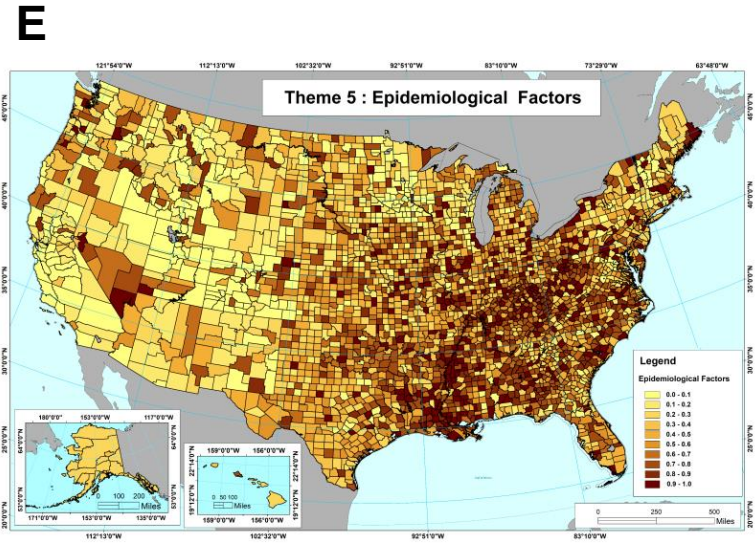
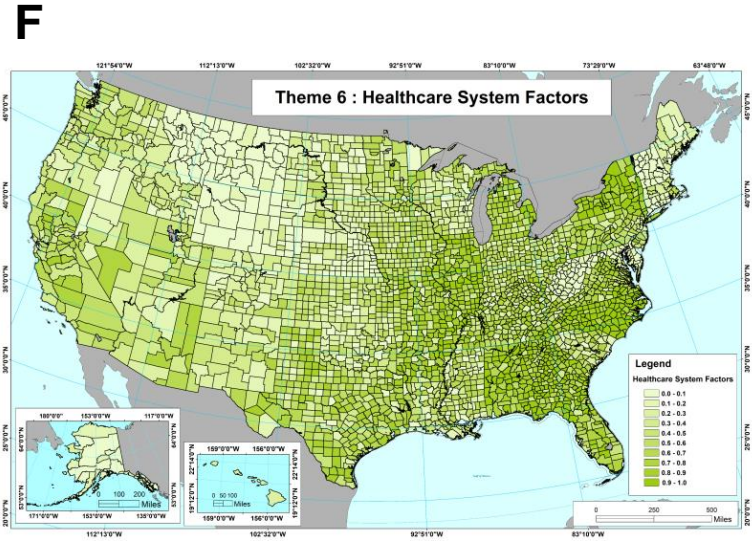

**Figure 2**

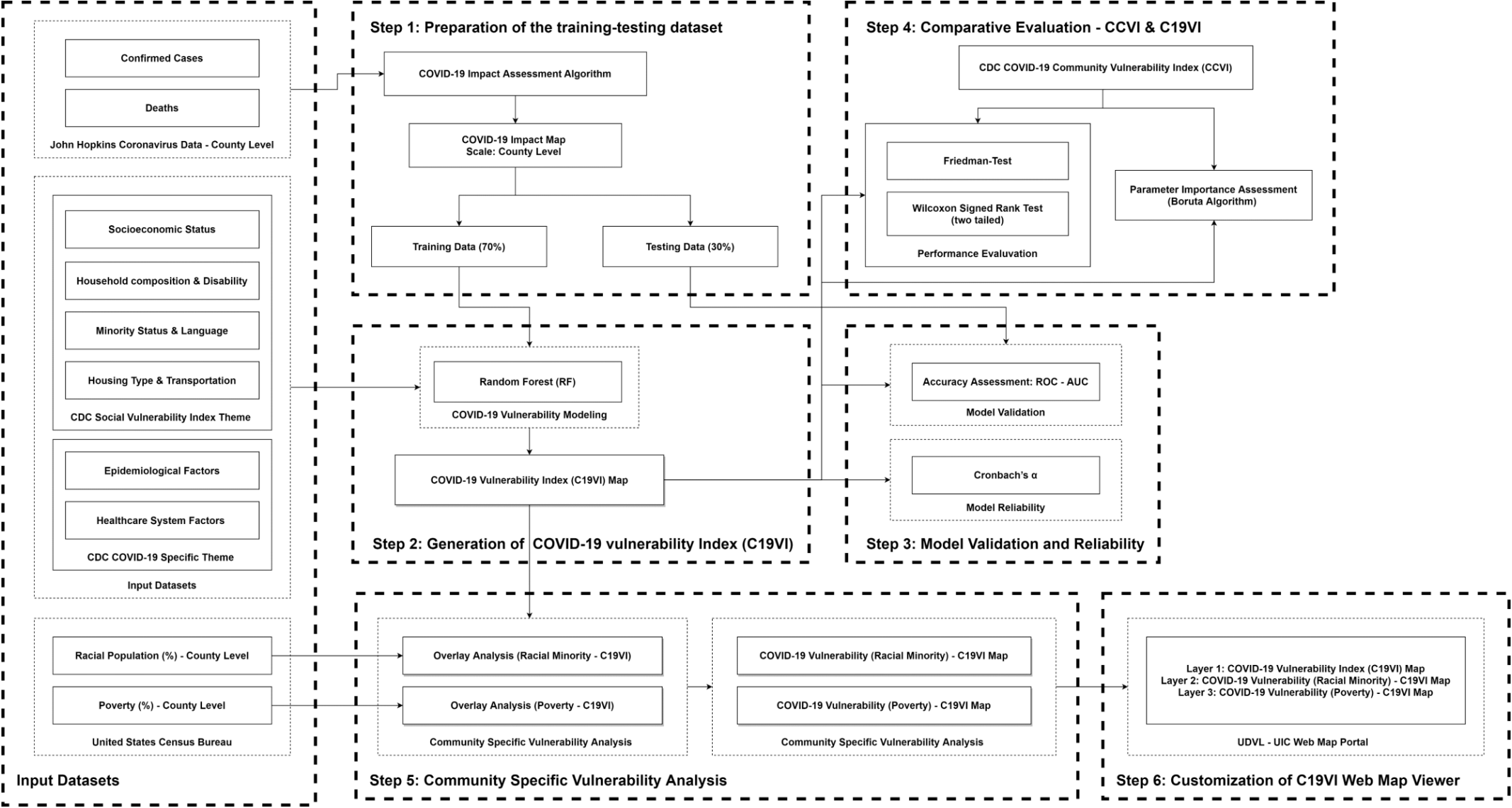

**Figure 3**

**Figure 4**

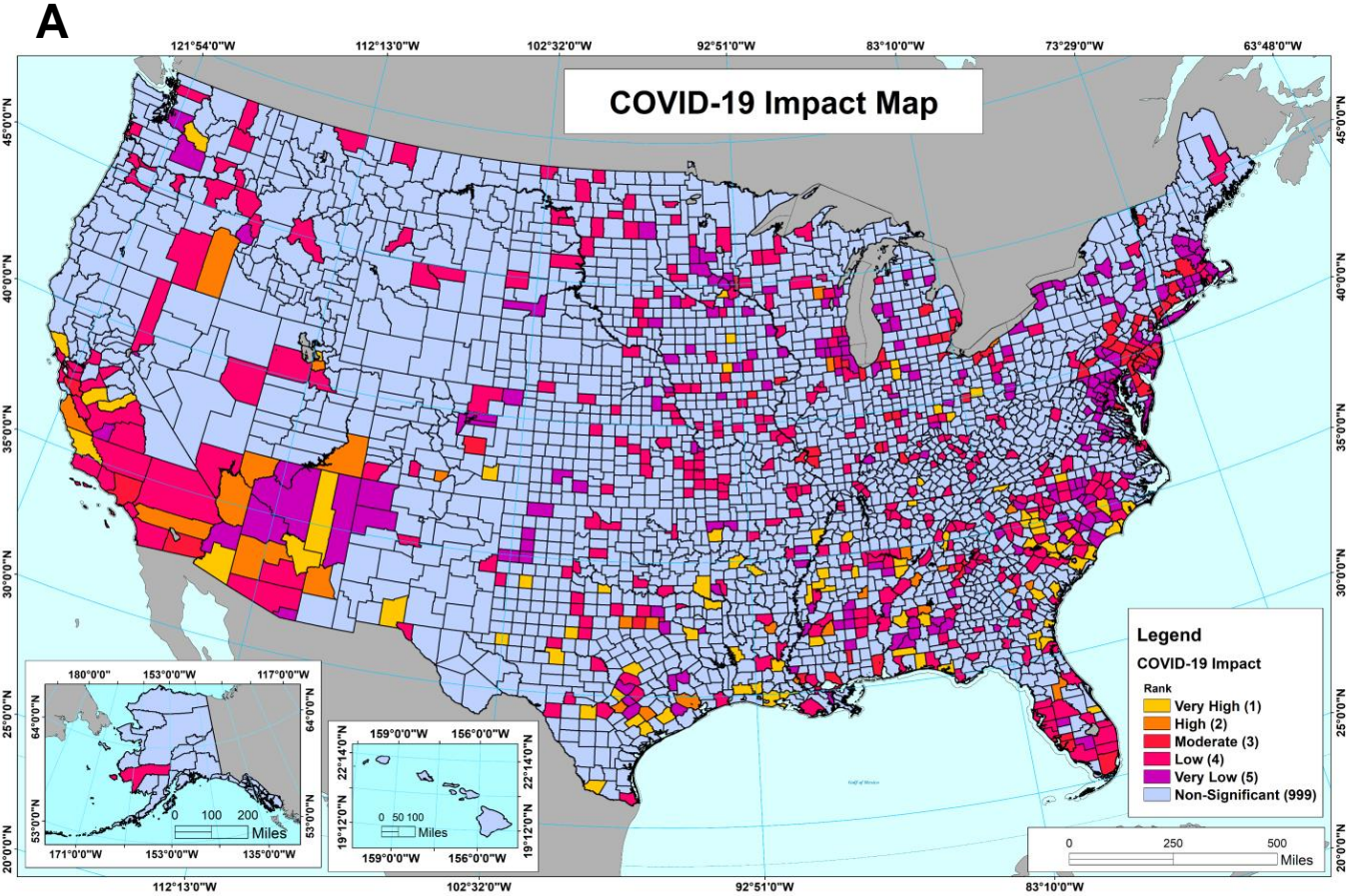 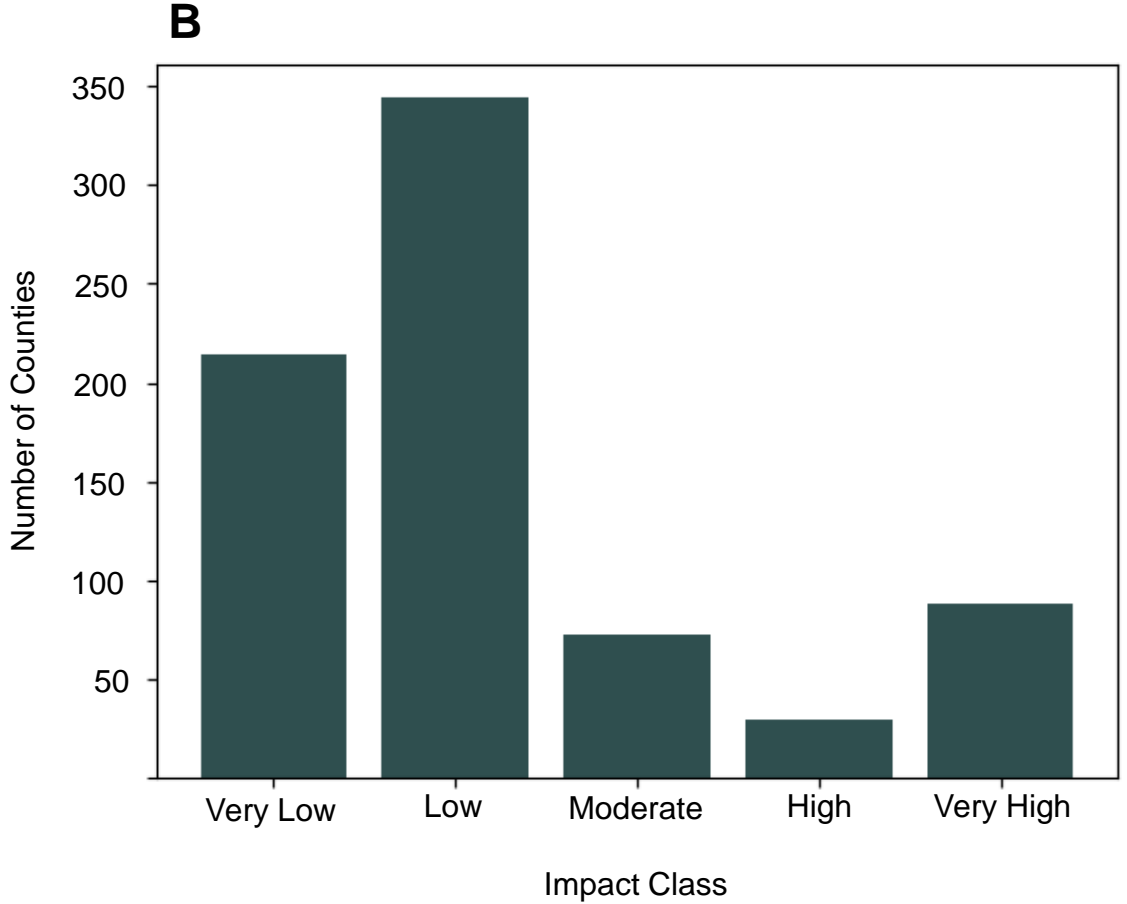

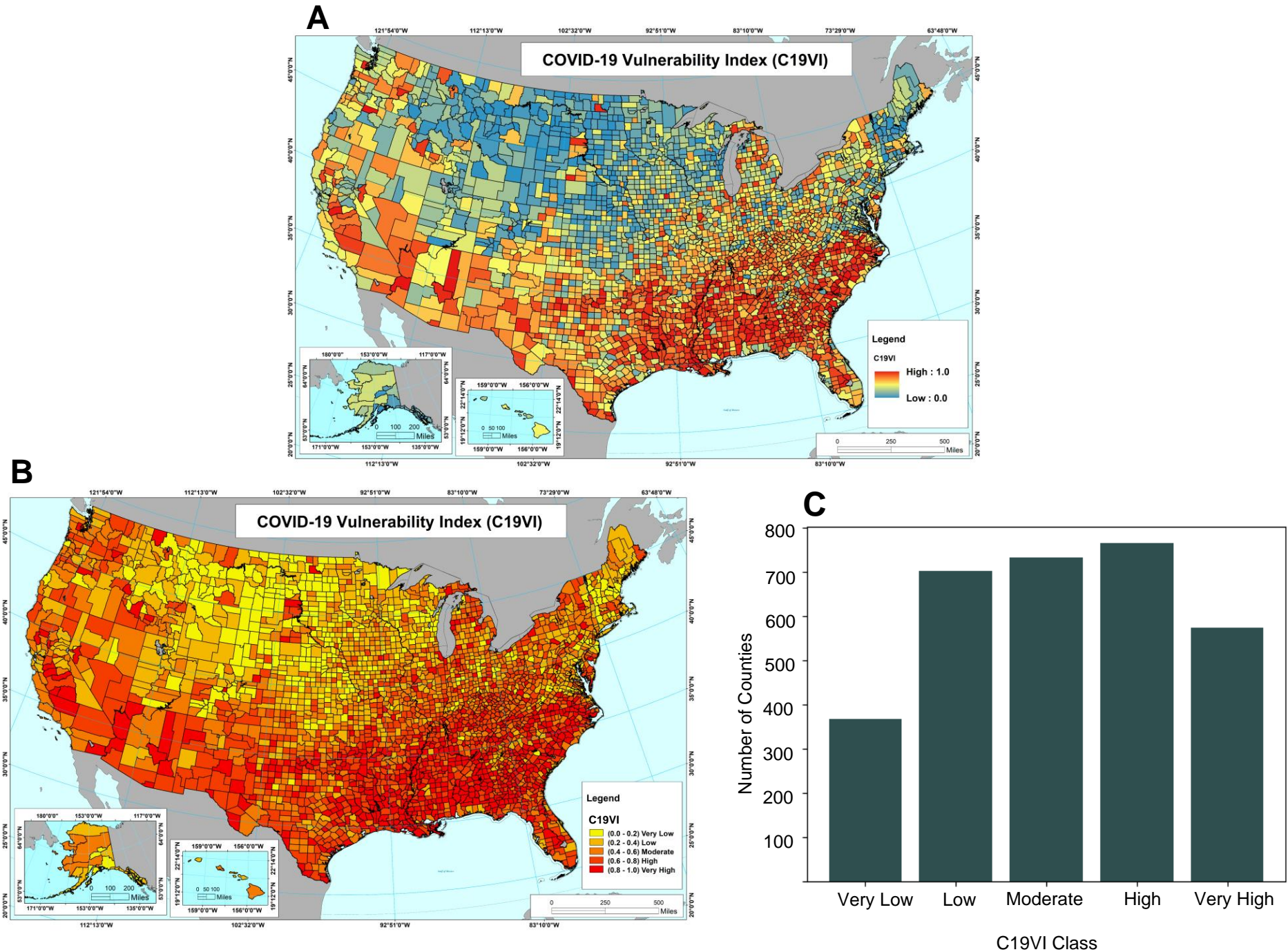

Figure 5

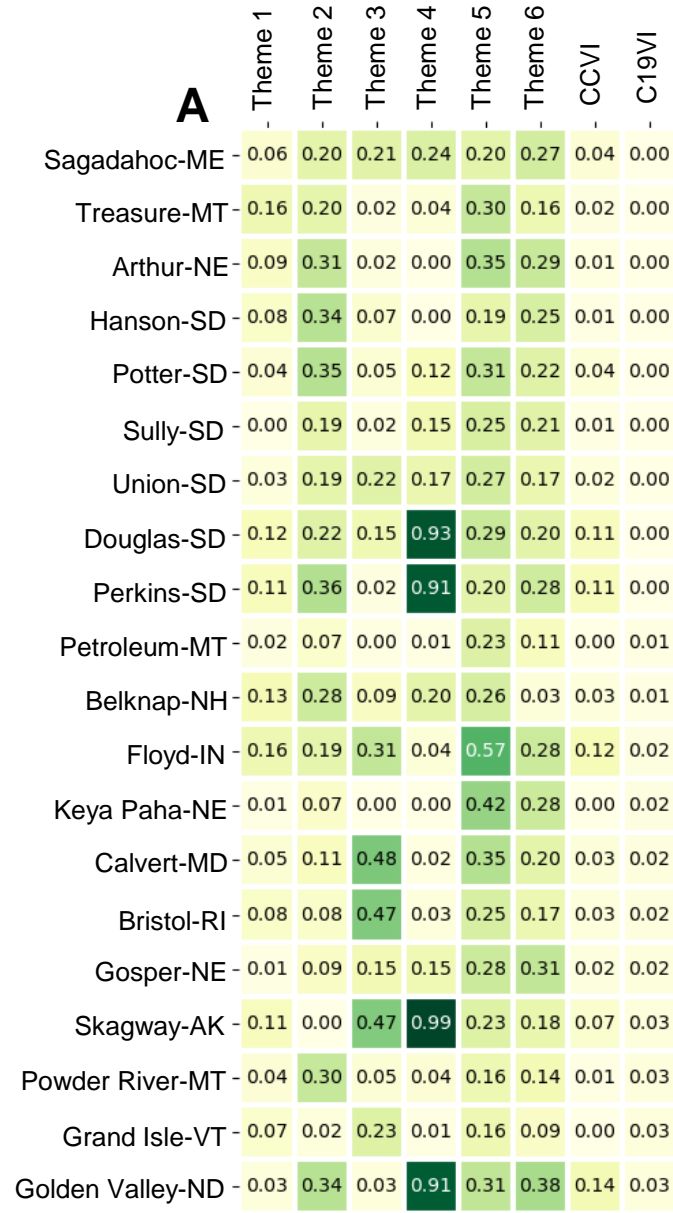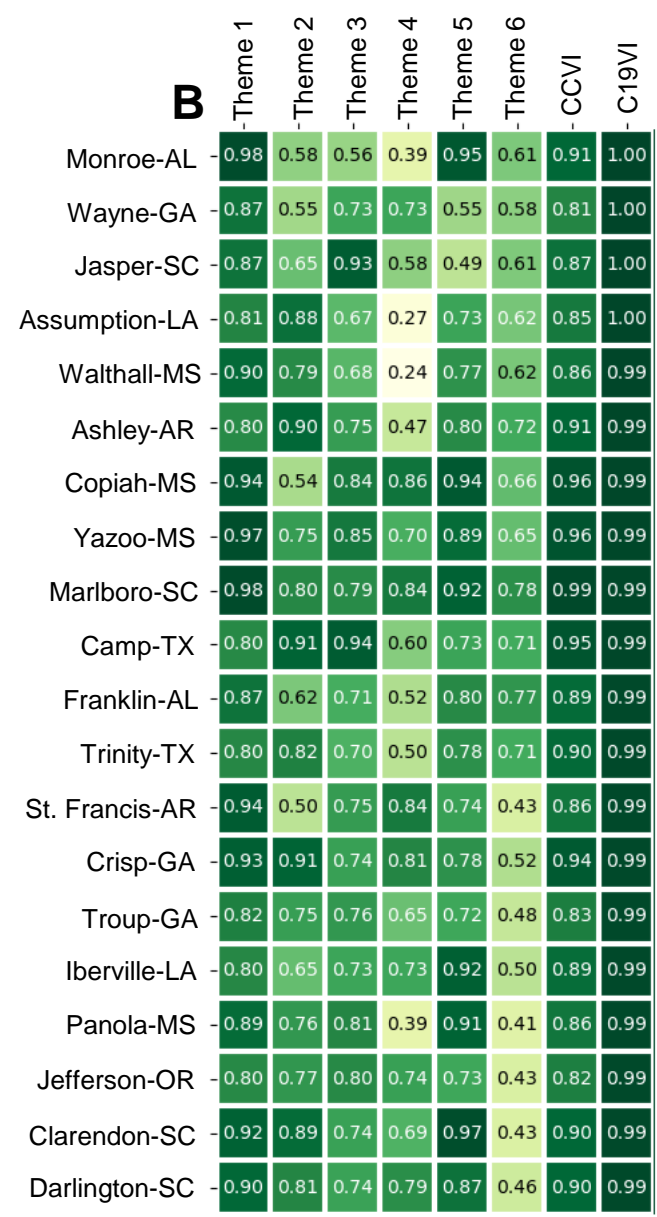

Figure 6

**Figure 7**

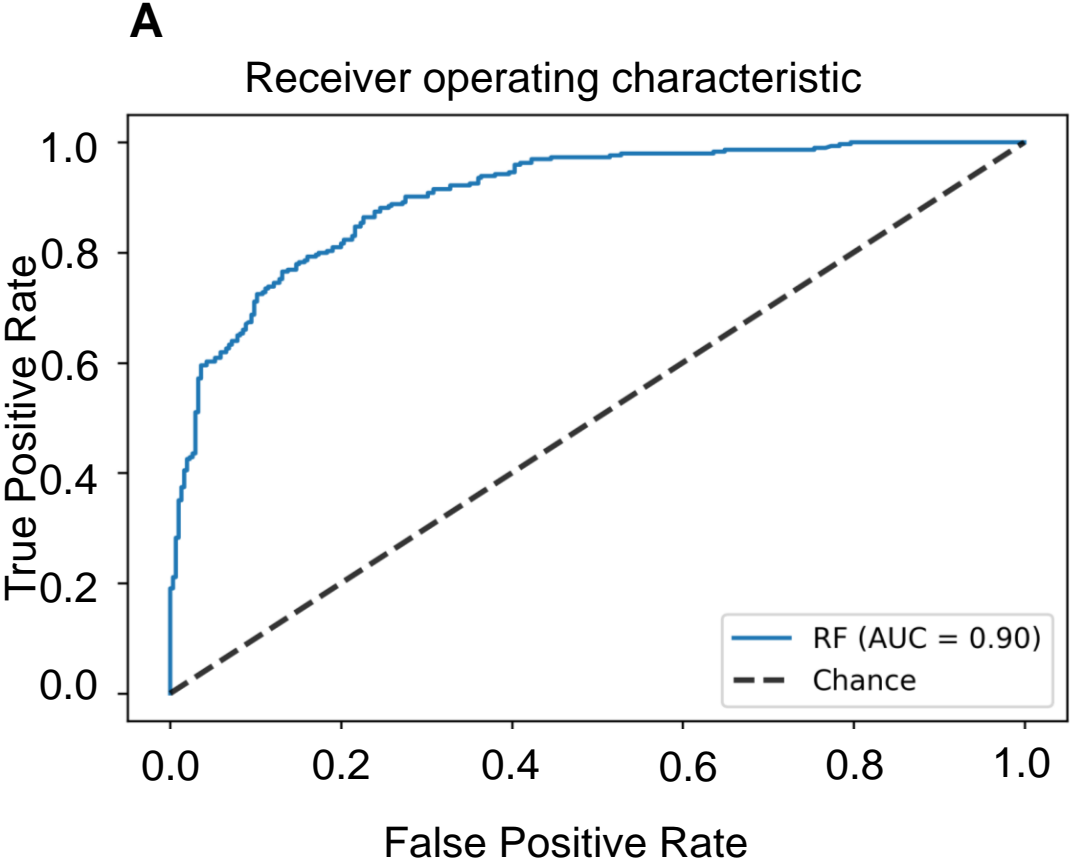 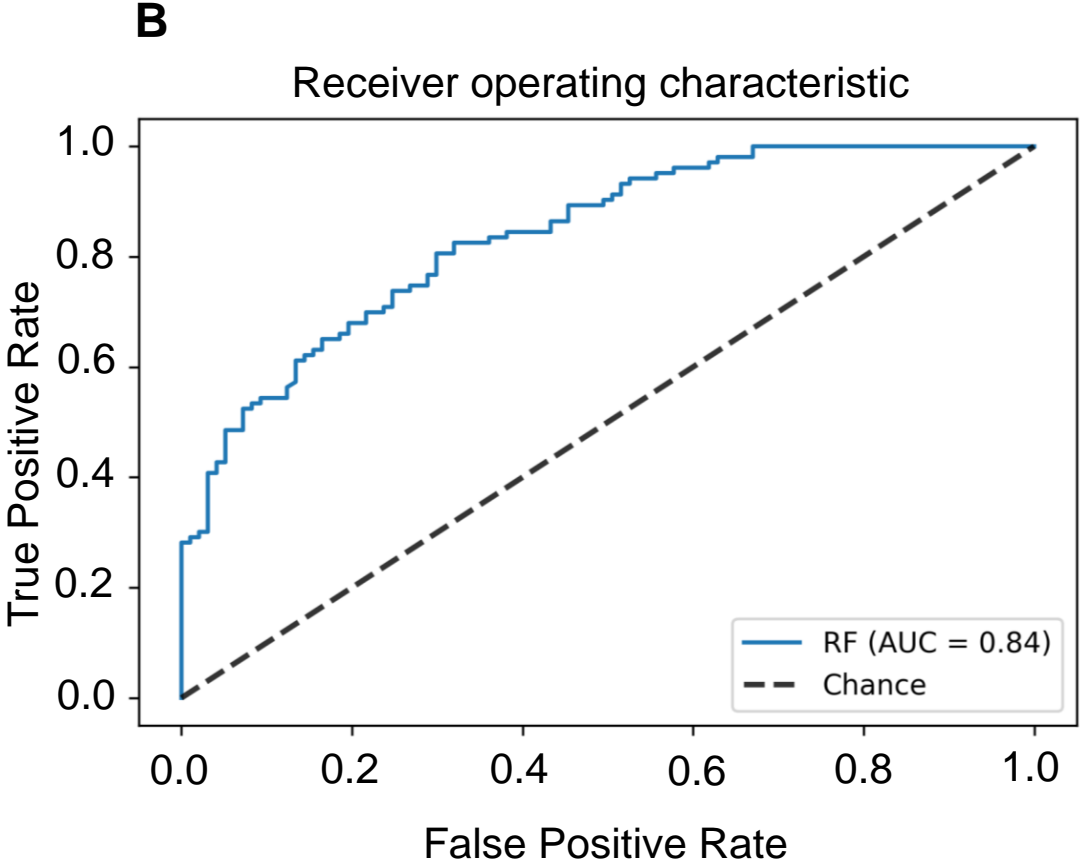

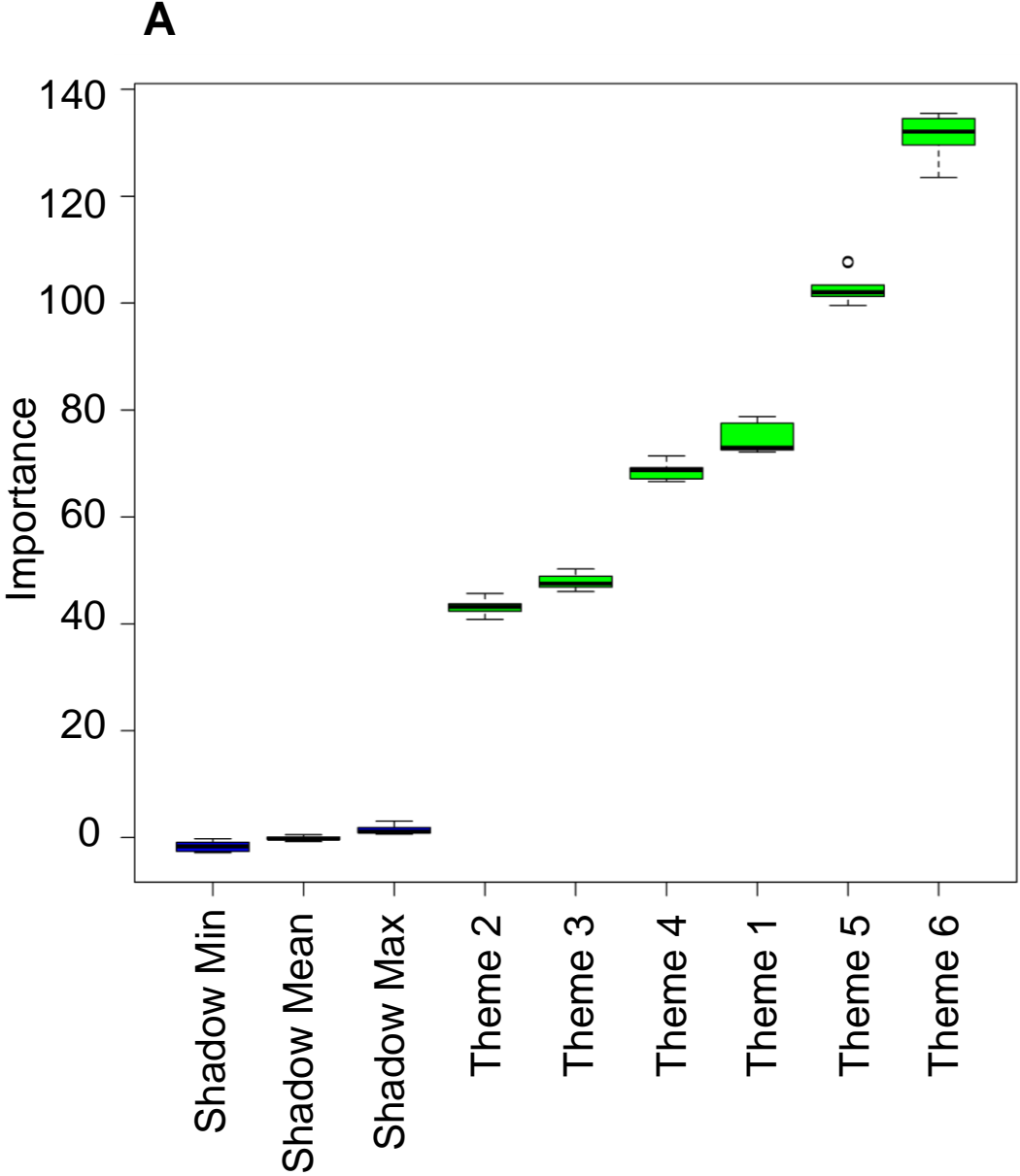 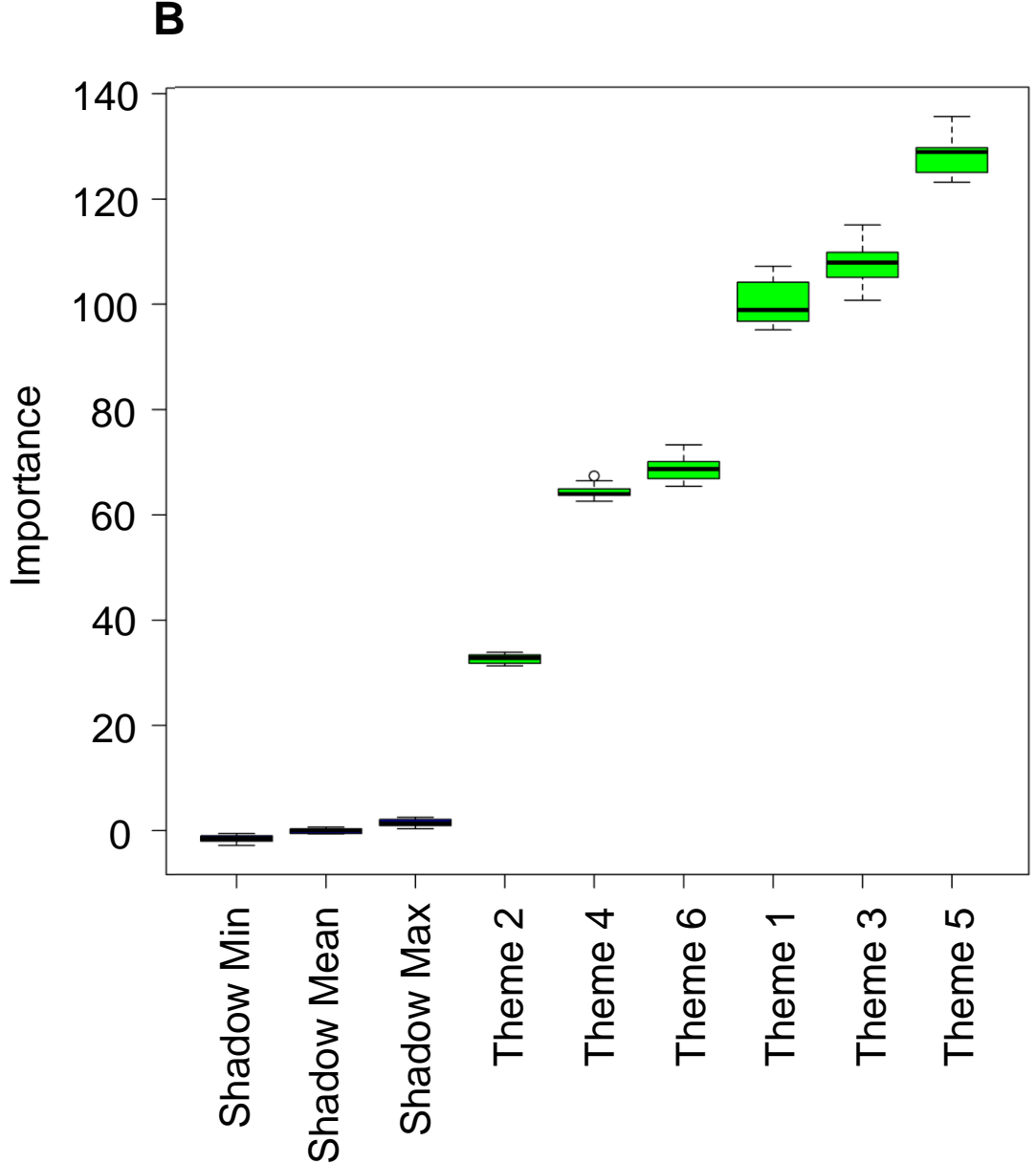

**Figure 8**

**Figure 9**

A 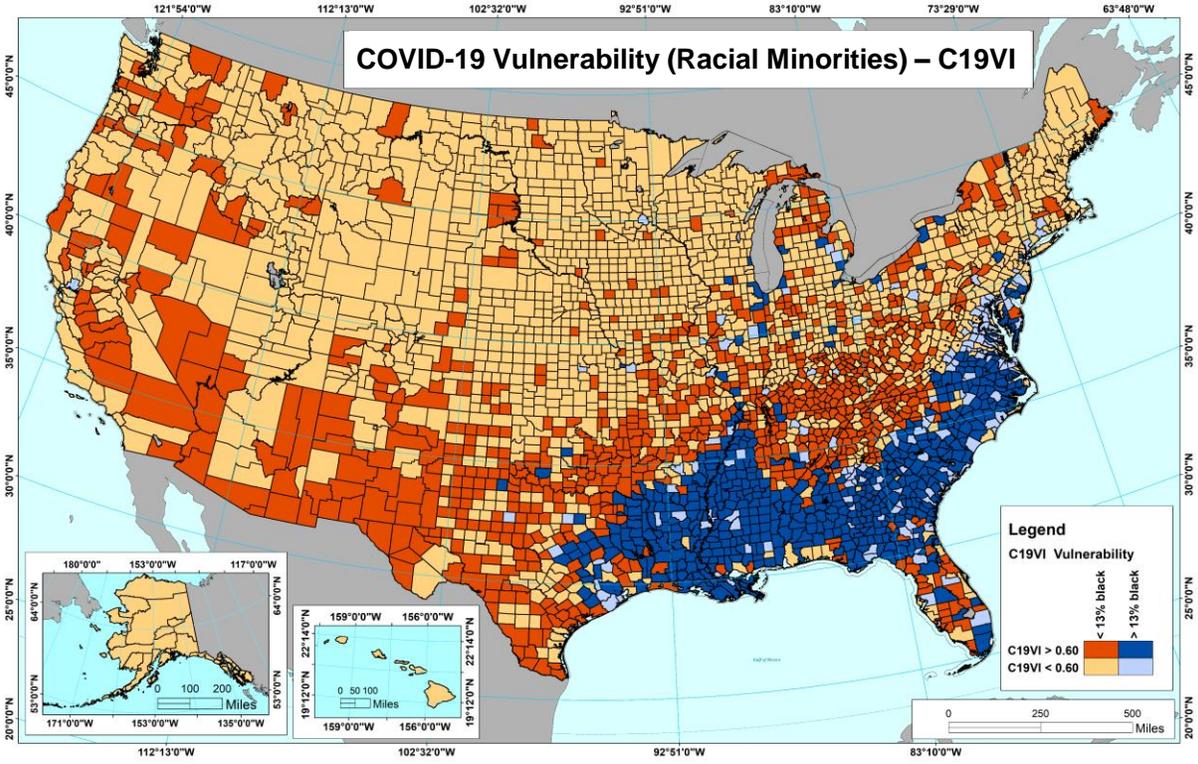

B 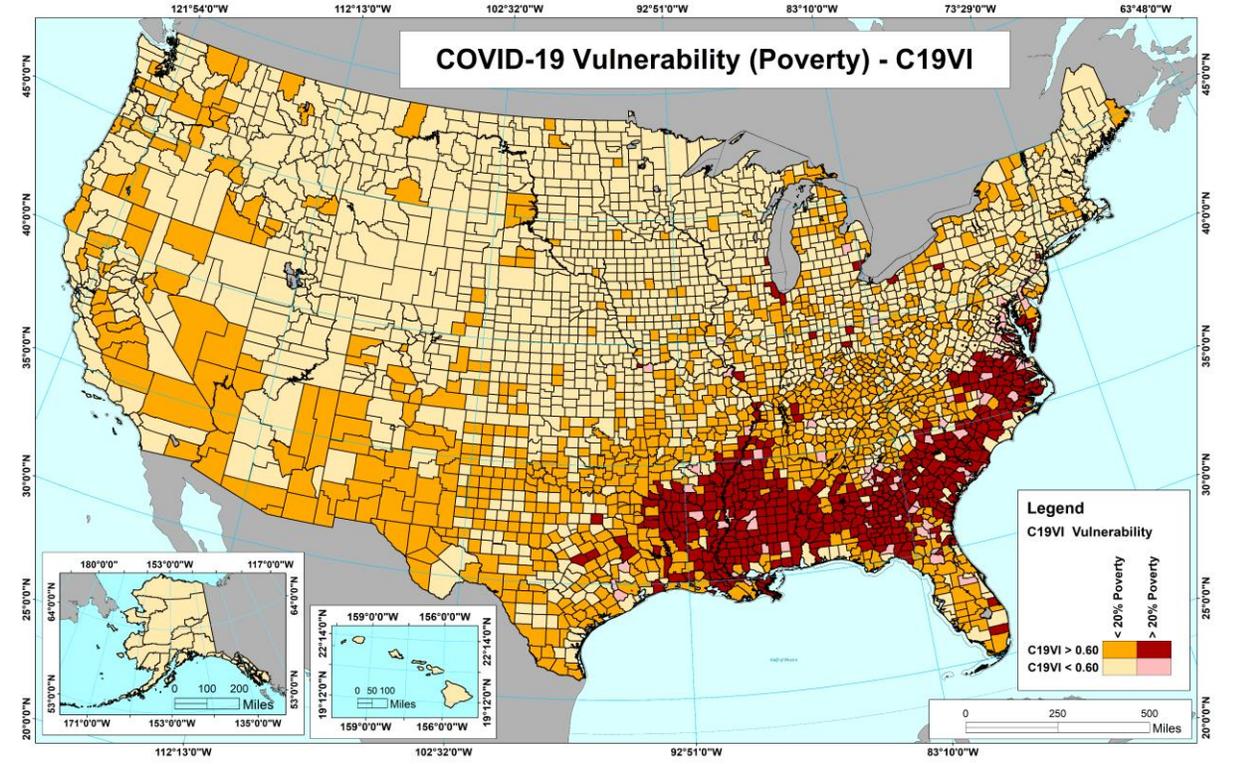

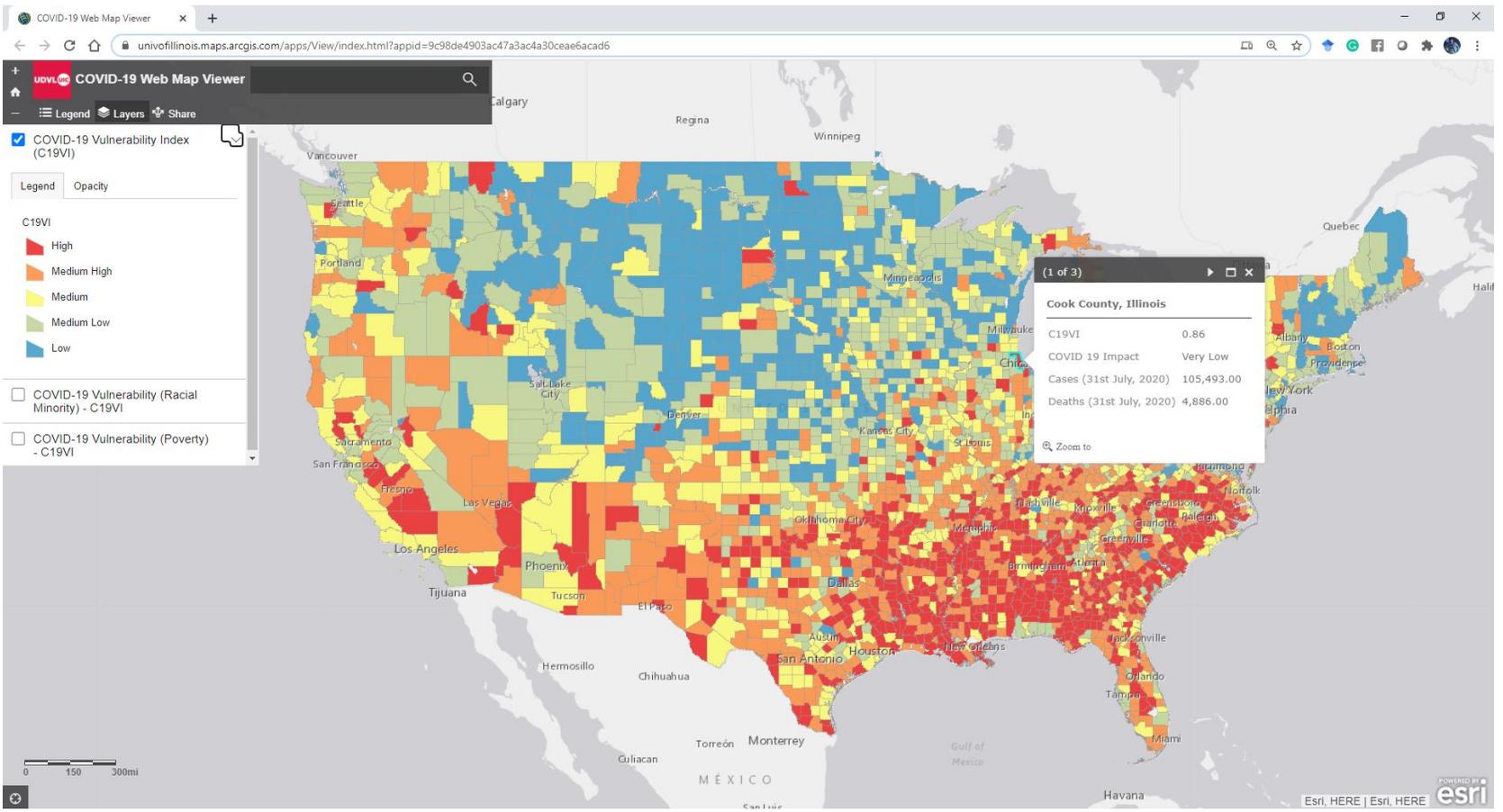
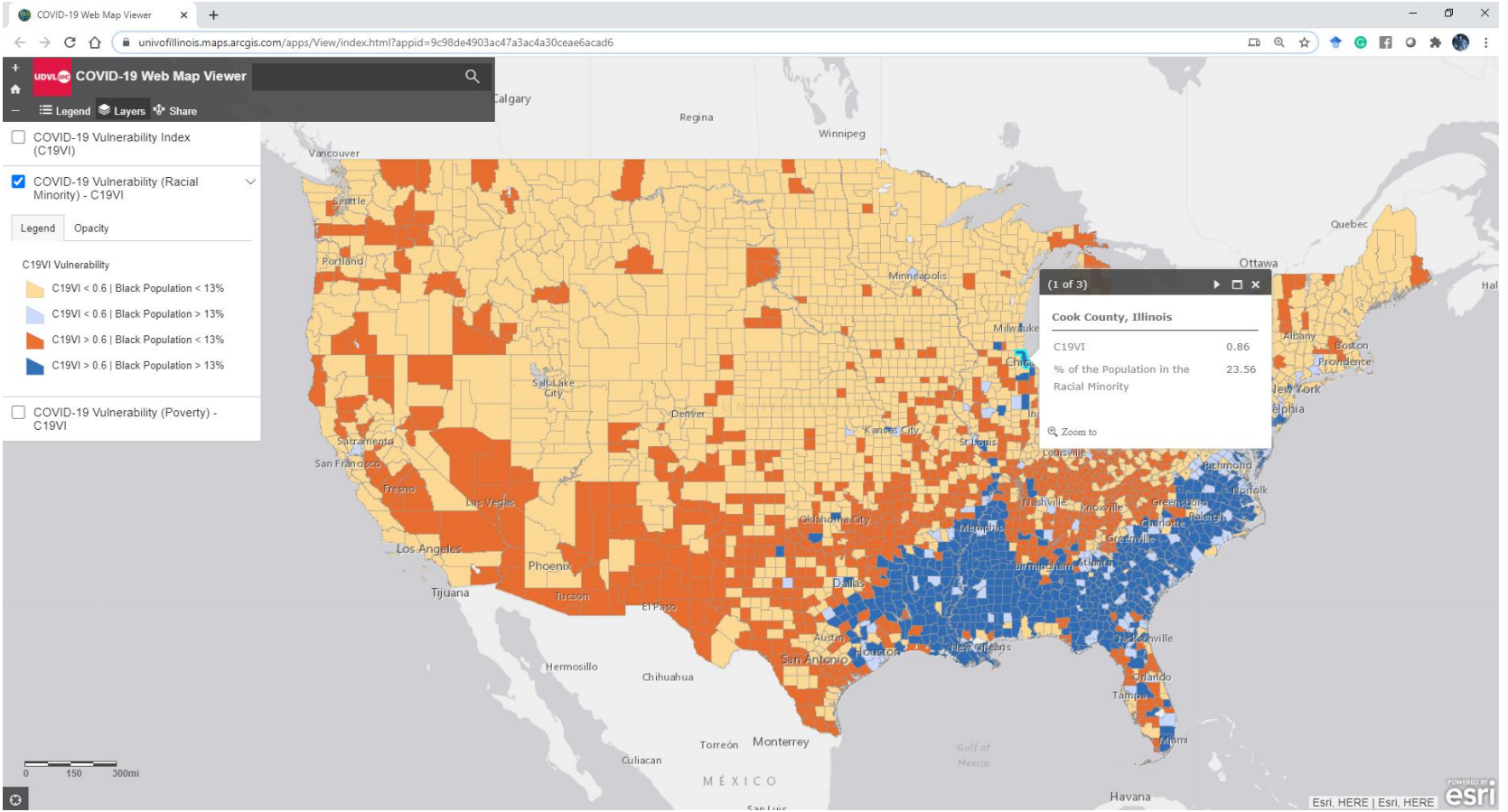
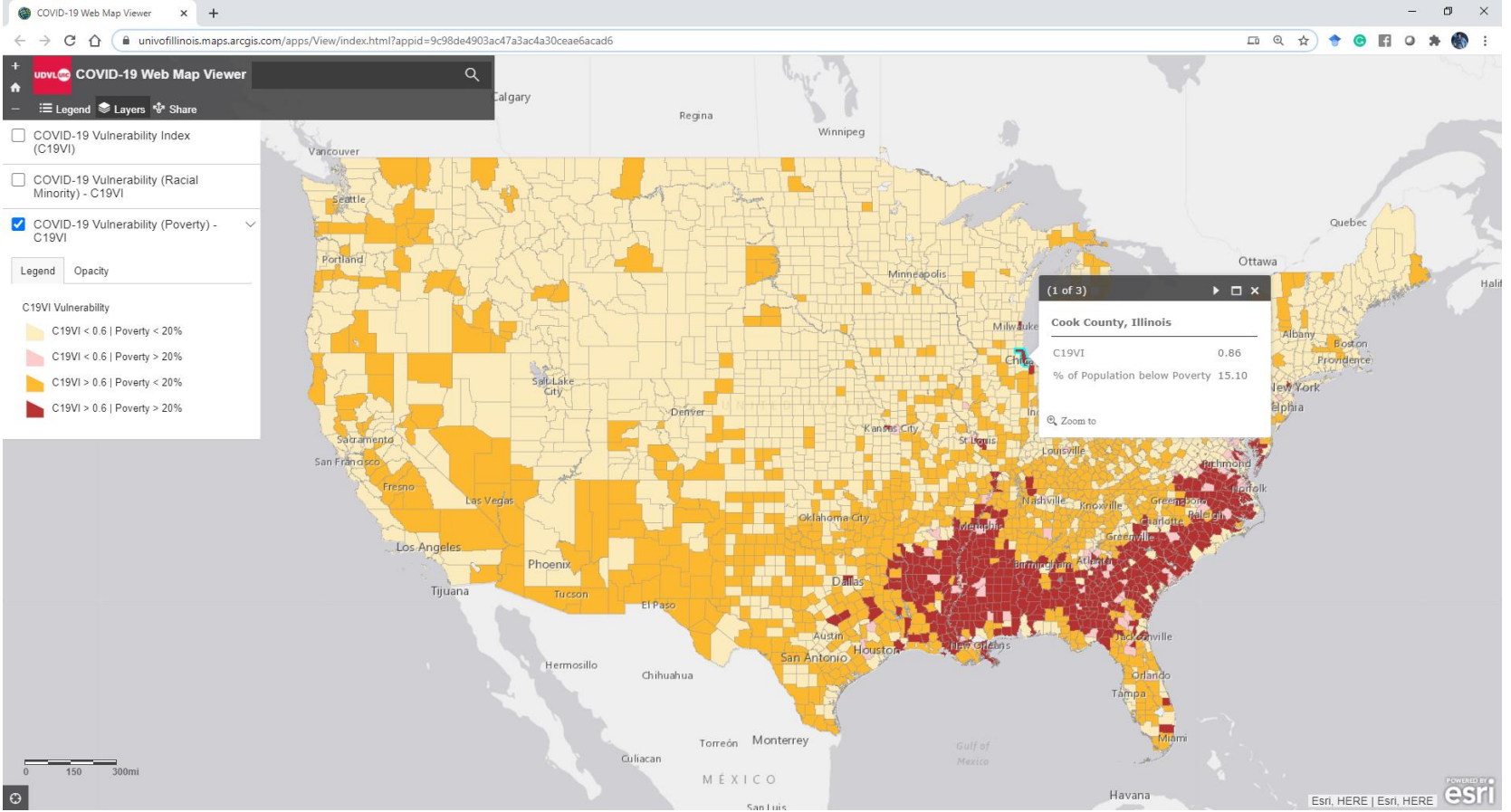

# Supplementary - 1
# COVID-19 Impact Assessment Algorithm - Pseudocode

---

**Algorithm 1:** COVID-19 Impact Assessment

**Input:** cases[][], deaths[][], n, c
    // cases[][] : Daily reported COVID-19 confirmed cases by county
    // deaths[][] : Daily reported COVID-19 deaths by county
    // n = 192; Total days from 22nd Jan 2020 to 31st July 2020
    // c: 3142; Total number of US counties

**Output:** countyImpactCOVID-19[][]
    // County wise COVID-19 impact (rank and score)

1 **Def** COVID-19Impact(*DType, Homogeneity, OverallT, Overalltt, preT, prett, postT, posttt*):
2     **if** *((Homogeneity == 'true') and (overallT == 'increasing'))* **then**
3         **Switch(DType):**
4             case('IFR'): Rank = 1; Score = overallG; **Break;**
5             case('D'): Rank = 2; Score = overallG; **Break;**
6             case('C'): Rank = 4; Score = overallG; **Break;**
7     **else**
8         **if** *((Homogeneity == 'false') and (preT == 'increasing'))* **then**
9             **if** *(postT == 'increasing')* **then**
10                 **Switch(DType):**
11                     case('IFR'): Rank = 1; Score = postG; **Break;**
12                     case('D'): Rank = 2; Score = postG; **Break;**
13                     case('C'): Rank = 4; Score = postG; **Break;**
14             **if** *(postT == 'notrend')* **then**
15                 **Switch(DType):**
16                     case('IFR'): Rank = 3; Score = preG; **Break;**
17                     case('D'): Rank = 3; Score = preG; **Break;**
18                     case('C'): Rank = 5; Score = preG; **Break;**
19             **if** *(postT == 'decreasing')* **then**
20                 Rank = 5; Score = - postG;
21         **else**
22             Rank = -999; Score = -999;
23     **return** Rank, Score
    // COVID impact rank and corresponding gradient (score)
24 **Def** TrendAnalysis(*TimeSeries*[]):
25     Trend = MannKendall(TimeSeries[])
26     Slope = SenSlope(TimeSeries[])
27     **return** Trend, Slope
    // Type of trend and corresponding gradient of trend
28 **Def** HomogeneityAnalysis(*TimeSeries*[]):
29     Homogeneity, ChangePoint = pettittTest(TimeSeries[])
30     **return** Homogeneity, ChangePoint
    // Homogeneity analysis and change point detection

```
31  Def Main():
32      for i = 1 to c do
33          for j = 1 to n do
34              C[j] = cases[i][j]           // Confirmed COVID-19 cases for a county
35              D[j] = deaths[i][j]          // COVID-19 deaths for a county
36              IFR[j] = deaths[i][j]/cases[i][j]  // IFR = deaths/confirmed cases
37          nziC = C[].index(next(filter(lambda x: x!=0, C[])))
38          nziD = D[].index(next(filter(lambda x: x!=0, D[])))
39          nziIFR = IFR[].index(next(filter(lambda x: x!=0, IFR[])))
                    // first non Zero Index of array or the instance when first
                    confirmed case, death and infection fatality rate reported
40          homoC, cpC = HomogeneityAnalysis(C[nziC:])
41          homoD, cpD = HomogeneityAnalysis(D[nziD:])
42          homoIFR, cpIFR = HomogeneityAnalysis(IFR[nziIFR:])
                            // Homogeneity analysis and change point detection
43          overallCT, overallCG = TrendAnalysis(C[nziC:])
44          preCT, preCG = TrendAnalysis(C[nziC, cpC-1])
45          postCT, postCG = TrendAnalysis(C[cpC:])
                    // Trend analysis overall, pre and post change point – cases
46          overallDT, overallDG = TrendAnalysis(D[nziD:])
47          preDT, preDG = TrendAnalysis(D[nziD, cpD-1])
48          postDT, postDG = TrendAnalysis(D[cpD:])
                    // Trend analysis overall, pre and post change point – deaths
49          overallIFRT, overallIFRG = TrendAnalysis(IFR[nziIFR:])
50          preIFRT, preIFRG = TrendAnalysis(IFR[nziIFR, cpIFR-1])
51          postIFRT, postIFRG = TrendAnalysis(IFR[cpIFR:])
                    // Trend analysis overall, pre and post change point - IFR
52          Rank-IFR, Score-IFR = COVIDImpact('IFR', homoIFR,
              overallIFRT, overallIFRG, preIFRT, preIFRG, postIFRT,
              postIFRG)
53          Rank-D, Score-D = COVIDImpact('D', homoD, overallDT,
              overallDG, preDT, preDG, postDT, postDG)
54          Rank-C, Score-C = COVIDImpact('C', homoC, overallCT,
              overallCG, preCT, preCG, postCT, postCG)
                // Priority wise COVID Impact assessment (IFR > deaths > cases)
55          rank = min(Rank-IFR, Rank-D, Rank-C)        // Best Rank
56          Switch(Rank):
57              case('Rank-IFR'): Score = Score-IFR; Break;
58              case('Rank-D'): Score = Score-D; Break;
59              case('Rank-C'): Score = Score-C; Break;
60          impactRank.insert(i, Rank)          // impactRank[] – Final rank
61          impactScore.insert(i, Score)        // impactScore[] – Final gradient
62      countyImpactCOVID19[][] = coulmnStack(impactRank[],
          impactScore[])
63      return countyImpactCOVID19[][]
```